\documentclass[lettersize,journal,10pt]{IEEEtran}

\pdfoutput=1
 
\usepackage{array}
\usepackage{textcomp}
\usepackage{stfloats}

\usepackage{tipa}
\usepackage{dsfont}
\usepackage{bbm}
\usepackage{color}
\usepackage{mathrsfs}
\usepackage{amsfonts}
\usepackage{cite,url,subfigure,epsfig,graphicx}%
\usepackage{amssymb,amsmath,bm,makecell}
\usepackage{indentfirst}
\usepackage{algorithmic}
\usepackage{algorithm}
\usepackage{epstopdf}
\usepackage{tabulary}
\usepackage{booktabs}
\IEEEoverridecommandlockouts
\usepackage{times,verbatim,amsfonts,amsmath,color}
\usepackage{amsthm}
\newtheorem{definition}{\textbf{Definition}}
\newtheorem{lemma}{\textbf{Lemma}}
\newtheorem{theorem}{\textbf{Theorem}}

\newtheorem{assumption}{\textbf{Assumption}}
\usepackage{soul}

\newcommand{\norm}[1]{\left\lVert#1\right\rVert}    
\usepackage{multirow}

\usepackage{hyperref}

\begin{document}

\title{Service Delay Minimization for Federated Learning over Mobile Devices
}
\author{
\IEEEauthorblockN{Rui Chen\authorrefmark{1}, Dian~Shi\authorrefmark{1}, Xiaoqi Qin\authorrefmark{2}, Dongjie Liu\authorrefmark{3}, Miao Pan\authorrefmark{1} and Shuguang Cui\authorrefmark{4}}
\IEEEauthorblockA{\authorrefmark{1}Department of Electrical and Computer Engineering, University of Houston, Houston, TX
77204}
\IEEEauthorblockA{\authorrefmark{2}School of Information and Communication Engineering, Beijing University of Posts and Tele., Beijing, China  100876}
\IEEEauthorblockA{\authorrefmark{3}Purple Mountain Laboratories, Nanjing, China 211189}
\IEEEauthorblockA{\authorrefmark{4}School of Science and Engineering, The Chinese University of Hong Kong, Shenzhen, China 518172}
}

\author{Rui Chen,~\IEEEmembership{Student Member,~IEEE},
        Dian Shi,~\IEEEmembership{Student Member,~IEEE},
        Xiaoqi~Qin,~\IEEEmembership{Member,~IEEE},
        Dongjie Liu,
        Miao Pan,~\IEEEmembership{Senior Member,~IEEE},
        and~Shuguang Cui,~\IEEEmembership{Fellow,~IEEE}
\IEEEcompsocitemizethanks{\IEEEcompsocthanksitem R. Chen, D. Shi, and M. Pan are with the Department of Electrical and Computer Engineering, University of Houston, Houston, TX, 77204 (e-mail: rchen19@uh.edu, dshi3@uh.edu, mpan2@uh.edu).
\IEEEcompsocthanksitem X. Qin is with the State Key Laboratory of Networking and Switching Technology, Beijing University of Posts and Telecommunications, Beijing 100876, P. R. China (e-mail: xiaoqiqin@bupt.edu.cn).
\IEEEcompsocthanksitem D. Liu is with Purple Mountain Laboratories, Nanjing, 211189, P. R. China (e-mail: liudongjie@pmlabs.com.cn).
\IEEEcompsocthanksitem S. Cui is with the School of Science and Engineering, the Chinese University of Hong Kong (Shenzhen), Shenzhen, P. R. China 518172 (e-mail: 	shuguangcui@cuhk.edu.cn).
}}

% make the title area
\maketitle
 
\IEEEdisplaynontitleabstractindextext
 
\IEEEpeerreviewmaketitle

\begin{abstract}
Federated learning (FL) over mobile devices has fostered numerous intriguing applications/services, many of which are delay-sensitive. In this paper, we propose a service delay efficient FL (SDEFL) scheme over mobile devices. Unlike traditional communication efficient FL, which regards wireless communications as the bottleneck, we find that under many situations, the local computing delay is comparable to the communication delay during the FL training process, given the development of high-speed wireless transmission techniques. Thus, the service delay in FL should be computing delay + communication delay over training rounds. To minimize the service delay of FL, simply reducing local computing/communication delay independently is not enough. The delay trade-off between local computing and wireless communications must be considered. Besides, we empirically study the impacts of local computing control and compression strategies (i.e., the number of local updates, weight quantization, and gradient quantization) on computing, communication and service delays. Based on those trade-off observation and empirical studies, we develop an optimization scheme to minimize the service delay of FL over heterogeneous devices. We establish testbeds and conduct extensive emulations/experiments to verify our theoretical analysis. The results show that SDEFL reduces notable service delay with a small accuracy drop compared to peer designs.

\end{abstract}


\section{Introduction}
With ever increasing computing and communication capabilities, beyond learning inference, mobile  devices (Google Pixel 4a with Adreno GPU, Mac with M1 chip and Wi-Fi 5, Nvidia Jetson devices with Wi-Fi 5, etc.) are promising to execute on-device training of some sophisticated deep learning (DL) models~\cite{CoreML}. Meanwhile, federated learning (FL) has evolved as a prospective distributed learning system across numerous mobile devices. FL enables mobile devices to learn DL models locally and then exchange the model updates via FL aggregation without data privacy leakage~\cite{mcmahan2017communication}. FL over mobile devices can provide a variety of services, including Google's next word prediction~\cite{hard2018federated}, vocal classifier~\cite{Siri}, cardiac event prediction~\cite{brisimi2018federated}, mobile object detection~\cite{posner2021federated}, etc. While many of those FL inspired services over mobile devices are delay-sensitive. FL service delays on mobile devices are primarily caused by two factors: local computing delays and wireless communication delays during training rounds. Because state-of-the-art DL models are typically overparameterized, it takes a long time to locally train those DL models on computational resource restricted mobile devices. In term of wireless communication delay, the size of gradients to exchange is relatively large (e.g., ResNet50 with ImageNet dataset has 98M parameters). This places a significant demand on wireless transmission and may result in huge aggregated communication delays during FL training. Such a service delay issue is more and more severe as the contemporary DL models become deeper and larger. As a result, it is vital to investigate how to develop a service delay efficient FL over mobile devices.

One of the most prominent ways for reducing computational complexity for on-device training is weight quantization~\cite{han2015deep, fu2021cpt, zhang2018lq,li2017training}. Weight quantization, by shrinking the precision of model parameters, may effectively reduce the storage sizes and the computing delay of on-device training. Large-scale deep neural networks (DNNs) (e.g., ResNet and MobileNet) might, for example, be quantized to provide a foundation for fast on-device training/inference, as described in~\cite{zhang2018lq,fu2021cpt}. Those approaches, on the other hand, strive to increase learning accuracy and primarily focus on centralized learning problem with a single device. The weight quantization's impacts on computing delay or service delay in the FL with multiple devices have not been fully explored yet. Since FL encompasses massive devices with heterogeneous computing capabilities and heterogeneous local datasets, it is worth further investigating the impacts of weight quantization on the service delay and FL learning performance.

For wireless transmission delay in FL, most existing works in wireless communities have mainly conducted the radio resource allocation under the FL convergence constraints~\cite{tran2019federated,nishio2019client, vu2020cell,chen2020joint}, while neglecting how to inherently reduce the communication payload and delay from the DL algorithms themselves. In contrast, assuming wireless transmissions constitute the bottleneck, two prominent solutions for improving FL communication efficiency have been presented in machine learning communities: (i) gradient quantization~\cite{ bernstein2018signsgd, alistarh2017qsgd}, which compresses the size of gradients to transmit and thus reduces communication bandwidth consumption per round in FL at the cost of precision; (ii) local computing control strategy, e.g., local stochastic gradient descent (SGD)~\cite{mcmahan2017communication}), which allows participating mobile devices to perform multiple training iterations locally before updating to the FL edge server, thereby avoiding communication after every local iteration and reducing the update frequencies. In addition to these advantages, an aggressive gradient quantization method may cause severe distortion on the gradients, necessitating extra communication rounds for compensation in order to achieve FL convergence\cite{bernstein2018signsgd}. Furthermore, increasing the number of local training iterations may result in large discrepancies across local models, resulting in a lower error-convergence in FL. Intensive local training also causes significant computing delay in FL. As a result, existing communication efficient FL designs~\cite{bernstein2018signsgd,alistarh2017qsgd,mcmahan2017communication}, which prefer aggressive compression strategy/intensive local training to reduce the communication bandwidth consumption per round/total number of communication rounds in FL, may not be optimal in terms of FL service delay over mobile devices. First, most of them apply the identical quantization strategy to all participants, ignoring the large computing latency and varied computing capabilities of those devices. Such designs cannot address the FL straggler issue and so slow down model aggregation. Second, they disregard the latest advancements in high-speed wireless transmissions.
 
For wireless transmission delay in FL, most existing works in wireless communities have mainly conducted the radio resource allocation under the FL convergence constraints~\cite{tran2019federated,nishio2019client, vu2020cell,chen2020joint}, while neglecting how to inherently reduce the communication payload and delay from the DL algorithms themselves. By contrast, in machine learning communities, assuming wireless transmissions are the bottleneck, there have been two mainstream strategies proposed to improve communication efficiency of FL: (i) gradient quantization~\cite{ bernstein2018signsgd, alistarh2017qsgd}, which compresses the size of gradients to transmit and thus reduces communication bandwidth consumption per round in FL at the cost of precision; (ii) local computing control strategy, e.g., local stochastic gradient descent (SGD)~\cite{mcmahan2017communication}), which allows participating mobile devices to perform multiple training iterations locally before updating to the FL edge server, thereby avoiding communication after every local iteration and reducing the update frequencies. In parallel with those benefits, an aggressive gradient quantization strategy may result in significant distortion on the gradients, and thus require more communication rounds for compensation to achieve FL convergence~\cite{bernstein2018signsgd}. Besides, increasing the number of local training iterations may lead to large discrepancies between local models, which consequently incurs an inferior error-convergence in FL. Intensive local training also introduces large computing delay in FL. Therefore, existing communication efficient FL designs~\cite{bernstein2018signsgd,alistarh2017qsgd,mcmahan2017communication}, which prefer aggressive compression strategy/intensive local training to reduce the communication bandwidth consumption per round/the total number of communication rounds in FL, may not be optimal w.r.t. the service delay of FL over mobile devices. First, most of them assign the identical quantization strategy to all the participants, and don't consider the involved huge computing delay and heterogeneous computing capabilities of those devices. Such designs cannot mitigate the FL straggler problem and slow down the model aggregation. Second, they ignore the recent advance of high-speed wireless transmissions.

By embracing the high-speed wireless transmission era (e.g., 5G, Wi-Fi 5, Wi-Fi 6 or 6G), the wireless communication bottleneck can be relieved for FL over mobile devices. Furthermore, we observe that local computing delay for on-device training is comparable to wireless communication delay, if the transmission rate is sufficiently high. For example, transmitting a ResNet20 model via 100 Mbps Wi-Fi 5 links takes 90ms, which is comparable to the time consumption of executing one-step local training on a mobile device with a modest GPU, e.g., 86ms for Jetson Xavier with 1.3TFLOPs on Cifar10 dataset. Thus, in order to optimize the service delay of FL over mobile devices, simply minimizing local computing/wireless transmission delay independently is not good enough, and the delay trade-off between ``working" (i.e., local computing) and ``talking" (i.e., wireless communications) has to be considered.

Motivated by the aforementioned observations and challenges, we develop in this paper a \textbf{S}ervice \textbf{D}elay \textbf{E}fficient FL (SEDFL) over mobile devices, with the design goal of minimizing the FL's service delay (local computing delay + wireless communication delay during FL training process) without sacrificing the learning performance. The proposed SDEFL jointly considers stochastic weight quantization and local computing control for efficient on-device local training, and gradient quantization for efficient local model update communications. Briefly, we empirically study the impacts of different quantization and control strategies on the learning performance and the service delay in FL. Based on the empirical observations and theoretical convergence analysis, we formulate the service delay minimization to determine the optimal quantization strategies and iteration number of local training with considering heterogeneous computing and communication conditions of mobile devices. We further establish testbed and conduct experiments to evaluate the proposed scheme and verify our findings. Our salient contributions are summarized as follows.
\begin{itemize}
\item We first empirically study how local computing control (i.e., deciding the number of local training iterations), weight quantization strategy, and gradient quantization strategy affect the service delay of FL over mobile devices, respectively. Insight from those studies is that the local computing control and gradient quantization play very important roles to determine the service delay. 

\item Based on our empirical observation, we develop an SDEFL scheme over mobile devices by jointly selecting the number of local training iterations, gradient quantization levels, and weight quantization levels. Here, mobile devices are allowed to quantize the model weights locally to speed up the computing delay and then quantize the local-update gradients to reduce the communication delay. Besides, we provide the theoretical analysis of the convergence upper bound of the proposed SDEFL scheme with flexible quantization strategies.

\item According to the derived theoretical convergence rate, we formulate the SDEFL problem as an integer nonlinear programming, where the number of local training iterations, weight quantization level, and gradient quantization level are decision variables. Geometric programming is exploited to develop feasible solutions.

\item We set up testbeds, and conduct extensive simulations and experiments to verify the effectiveness of our proposed SDEFL scheme under various learning models, different data distributions across devices, and multiple wireless environmental settings. 
\end{itemize}
 
The rest of this paper is organized as follows. In Section~\ref{Sec:SysM}, the system model and problem setting are introduced. The empirical studies are discussed in Sections~\ref{Sec:SysM1}. The SDEFL's formulation, convergence analysis and feasible solutions are illustrated in Sections~\ref{Sec:CovAls}. In Section \ref{Sec:Result}, the experimental and simulation results are presented and analyzed. The related work is provided in Section \ref{sec:ReWork} and the paper is concluded in Section \ref{Sec:con}.

\section{Federated learning with Weight and Gradient Quantizations}\label{Sec:SysM}

\subsection{FL framework with weight and gradient quantization}
\begin{figure} \centering
  %{\includegraphics[width=\linewidth]{System.eps}}
  {\includegraphics[width=0.9\linewidth]{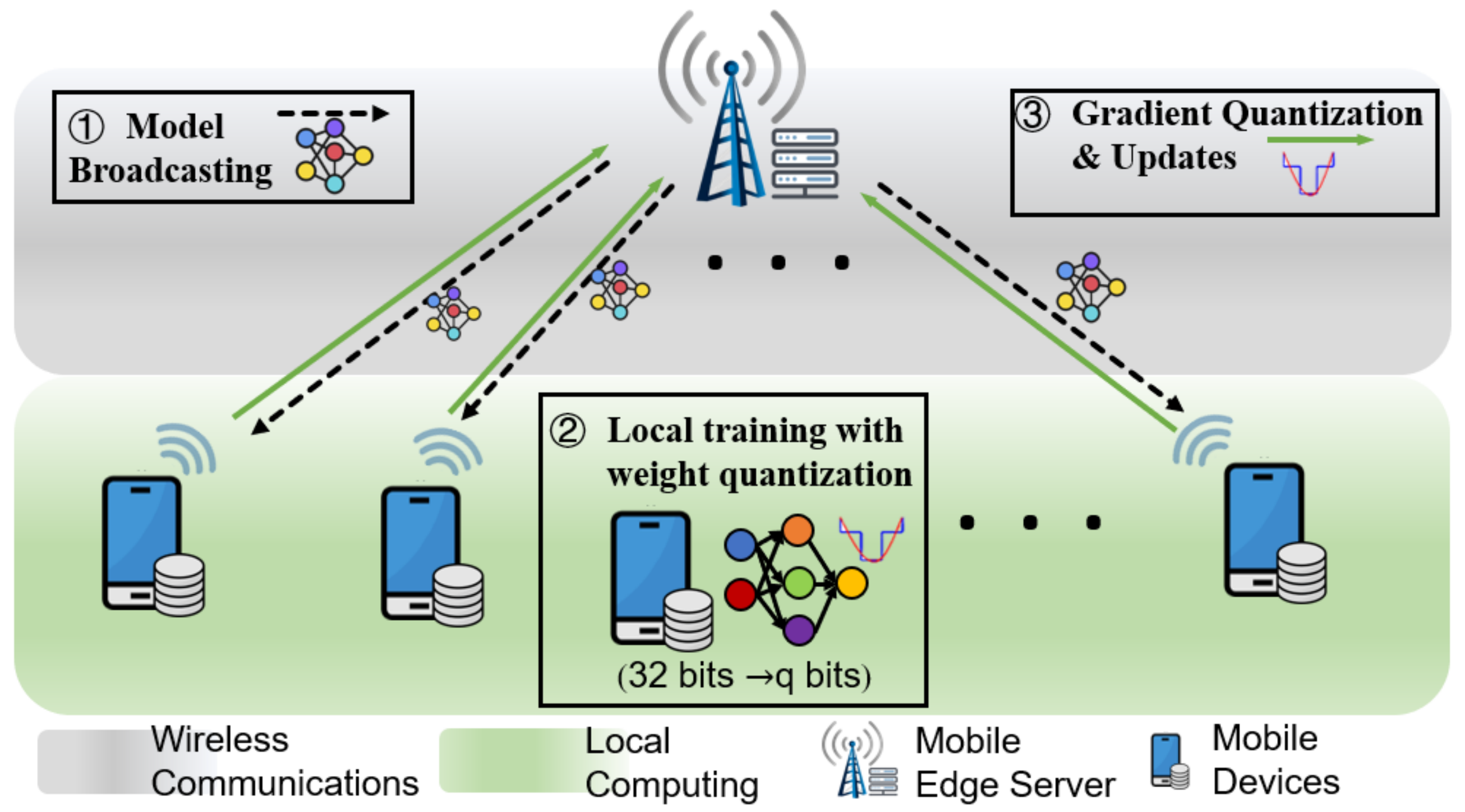}}
 \caption{Federated edge learning system with quantization.} \label{fig:System}
 \vspace{-0.2in}
\end{figure}

We consider a wireless FL system consisting of one mobile edge server (e.g., base station or gNodeB) and a set $\mathcal{N} = \{1,2,\cdots,N\}$ of mobile devices, shown in Fig~\ref{fig:System}. Each device $n$ has its private dataset $\mathcal{D}_n$ with $D_n = |\mathcal{D}_n|$ training data samples. All devices collaboratively train a global DNN model under the coordination of an mobile edge server\footnote{For simplicity, this paper only considers synchronous FL settings. We left the asynchronous FL setting as future work.}. The goal is to find the optimal model $\boldsymbol {w}\in \mathbb{R}^d$ to minimize a global empirical risk $ F(\boldsymbol {w})$ as follows
\begin{equation}
    \min_{\boldsymbol {w} \in \mathbb{R}^d} F(\boldsymbol {w}) =  \sum_{n=1}^N p_n F_n(\boldsymbol {w}), \label{FL_obj}
\end{equation}
where $d$ denotes the total number of model parameters and $p_n$ is the weight of device $n$ such that $\sum_{n=1}^N p_n =1$. $F_n(\boldsymbol {w}) = \frac{1}{D_n} \sum _{i\in \mathcal{D }_n }f_n^i(\boldsymbol {w};x_n^i,y_n^i)$ is the local training loss of device $n$, where $(x_n^i,y_n^i)$ is the $i$-th training sample in $\mathcal{D}_n$. Thus, we have $ p_n = {D_n}/{\sum_{n=1}^N D_n}$. 

To achieve high efficient FL over mobile devices, in this paper, we propose the FL with \underline{W}eight and \underline{G}radient \underline{Qu}antization (FL-QuWG) algorithm by considering multi-facet techniques, including local computing control (i.e., $H$), gradient quantization (i.e., $q_{g}$), and on-device weight quantization (i.e., $q_{w}$) during local model updates. Local computing control allows mobile devices to perform multiple local training iterations between every two sequential global aggregations, which reduces the total number of FL communication rounds. Gradient quantization enables devices to reduce the transmission bits of the updated gradients. Weight quantization allows mobile devices to store the model parameters and conduct local training with low-precision values to save the memory access and computing delay. 

Despite that weight quantizators and gradient quantizators are applied at different FL training stages, their mathematical definitions are the same. Here, we consider an unbiased stochastic quantization scheme, defined as follows.
%, e.g., 16-bit, 8-bit fixed-point numbers
\begin{definition}[Unbiased Quantization Scheme \cite{li2017training}]\label{def:q}
A randomized mapping $Q: \mathbb{R}^d \rightarrow \mathbb{R}^d$ is an \textit{unbiased quantization scheme} if there exists $\delta$ such that $\mathbb{E}[Q(\boldsymbol {w})] = \boldsymbol {w},$ $\mathbb{E}\left[\norm{Q(\boldsymbol {w})-\boldsymbol {w}}_2^2\right] \leq \delta\norm{\boldsymbol {w}}_2^2, \forall \boldsymbol {w} \in \mathbb{R}^d.$
\end{definition}
In general, each element in $\boldsymbol {w} \in \mathbb{R}^d$ will be independently quantized between $\arg \min \boldsymbol {w}$ and $\arg\max \boldsymbol {w}$. Then $\delta$ can be defined as $\delta = ({1+\sqrt{2d-1}})/{2(2^{q}-1)}$~\cite{li2017training}. We note that a smaller $q$ leads to a higher variance (i.e., a larger $\delta$)\footnote{We observe that commonly used quantization schemes stochastic rotated quantization scheme~\cite{basu2019qsparse} share similar trend and hence we consider the stochastic quantization scheme as an example in this paper to analyze the effect of quantization due to the space limit.}. We denote the coefficients $\delta_g$ and $\delta_w$ for gradient quantization and weight quantization schemes, respectively.

The pseudocode of the FL-QuWG algorithm is presented in Alg. \ref{alg:OverallFramwork}. Mathematically, all mobile devices start at the same initial point ${\boldsymbol {w}}_0$. Each device $n$ first quantizes the global model $\boldsymbol {w}$ with $q_{w,n}$ bits ($q_{w,n} \leq 32$) locally and runs $H$ local training iterations with quantized weights and their local datasets. After completing local on-device training, each device $n$ compresses the updated gradients with $q_{g,n}$ bits ($q_{g,n} \leq 32$) and transmits the quantized gradients to the edge server. The edge server then aggregates the gradient information and broadcasts the updated model to the mobile devices. The above procedure iterates until FL terminates. Let $K$ be the total training iterations, and then the total number of FL rounds is $K/H$.

\subsection{{The Service delay of FL}}
Given the training procedure in Alg.~\ref{alg:OverallFramwork}, the service delay of FL involves computing delay for local training (Line 4-5) and wireless communication delay (Line 7) in each round\footnote{We neglect the costs of the model broadcasting and the model aggregation (Line 10), which are taken by the powerful FL server and whose delay is much smaller than that of model updates.}. We assume the communication/computing delay of a mobile device for every round during FL training is the same, but varies among devices due to device heterogeneity. Therefore, we can neglect the indices for the FL communication rounds. Let $T_n$ denote the per-round service delay of device $n$ to perform on-device training and model updates. We have
\begin{flalign}
       T_n \triangleq T_n^{cp} (q_{w,n}, H) + T_n^{cm}(q_{g,n}), 
\end{flalign}
where $T^{cp}(q_{w,n}, H)$ denotes the \underline{c}om\underline{p}uting delay related to weight quantization $q_{w,n}$ and local computing control $H$, and $T^{cm}(q_{g,n})$ represents the \underline{c}om\underline{m}unication delay per FL round with gradient quantization $q_{g,n}$. Given straggler issues in synchronized FL systems, the per-round service delay $T$ is determined by the slowest participant, i.e., $T = \max_{n \in \mathcal{N}} \{T_n\}.$ Suppose the FL model converges in $K$ iterations using Alg.~\ref{alg:OverallFramwork}, and the corresponding service delay is 
\begin{flalign}
       T_{tot} \triangleq \frac{K}{H} \cdot \max_{n \in \mathcal{N}} \quad \{T_n\}. \label{servie_delay}
\end{flalign}

Our objective is to minimize the service delay of FL over mobile devices while guaranteeing FL convergence. We employ the expected gradient norm as an indicator of convergence~\cite{bottou2018optimization} due to the non-convexity of $F(\cdot)$, where the algorithm achieves an $\epsilon$-suboptimal solution, if {\small$\mathbb{E}\left[\frac{1}{K} \sum_{k=0}^{K-1} \mathbb{E}\left[\norm{\triangledown{F}(\boldsymbol {w}^{k})}_2^2\right]\right]\leq \epsilon$}. When $\epsilon$ is arbitrarily small, this condition can guarantee the algorithm converges to a stationary point after $K$ training iterations.
 
Thus, when the FL-QuWG achieves an $\epsilon$-suboptimal solution in $K$ training iterations, the service delay $T_{tot}$ is
\begin{flalign}
      \min &\quad  T_{tot} \label{eq:time_obj}\\
      \text{s.t.,} &\quad \mathbb{E}\left[\frac{1}{K} \sum_{k=0}^{K-1} \mathbb{E}\left[\norm{\triangledown{F}(\boldsymbol {w}^{k})}_2^2\right]\right]\leq \epsilon. \label{eq:convergence_ori}
\end{flalign}

Next, we conduct a series of empirical studies to analyze how service delay in (\ref{servie_delay}) and FL convergence in (\ref{eq:convergence_ori}) are affected by different strategies (i.e., $H$, $q_{w,n}$, $q_{g,n}$).

%%%%%%%%%%%%%%%%%%
\begin{algorithm}[t]
\caption{FL with Weight and Gradient Quantization
%Service Delay Efficient Federated Learning (SDEFL) 
} 
\label{alg:OverallFramwork} 
\begin{algorithmic}[1]  
\REQUIRE initial point $\boldsymbol{w}^{0} =\boldsymbol{w}_n^{0}$, learning rate $\eta$, number of training iterations $K$ \\
\ENSURE $\boldsymbol{w}^{K}$
\FOR{$k = 0,\cdots,K-1$}
%\STATE The FL server broadcasts $\boldsymbol{w}^k$ to devices $n \in \mathcal{N}$.mini-batch size $M$,
\FOR{mobile device $n \in \mathcal{N}$ in parallel}
\STATE Set ${\boldsymbol{w}}_n^{k+1} \leftarrow \boldsymbol{w}_n^{k+1-H} - {\boldsymbol{\Delta}}_{Q}^{k+1-H}$ 
\STATE Computing stochastic gradient \\
$\triangledown \widetilde{f}_n \left({\boldsymbol{w}}_n^{k}\right) = \triangledown {f}_n\left({\boldsymbol{w}}_n^{k}, \widetilde{{x}}_n,\widetilde{{y}}_n \right)$ for $\widetilde{{x}}_n,\widetilde{{y}}_n \in \mathcal{D}_n$
\STATE Update the model parameters\\
$ {\boldsymbol{w}}_n^{k+1} \leftarrow Q_{w,n}\left( {\boldsymbol{w}}_n^{k} - \eta_n \triangledown \widetilde{f}_n \left({\boldsymbol{w}}_n^{k}\right)\right)$
\IF{$((k+1) \mod H) = 0 $} 
\STATE Send $Q_{g,n}(\boldsymbol{w}_n^{k+1-H} - {\boldsymbol{w}}_n^{k+1})$ to the FL server
\ENDIF
\ENDFOR

\STATE The FL server aggregates \\
$ {\boldsymbol{\Delta}}_{Q}^{k+1-H} \leftarrow  \sum_{n=1}^N p_n Q_{g,n}(\boldsymbol{w}_n^{k+1-H} - {\boldsymbol{w}}_n^{k+1})$ 
\ENDFOR
\end{algorithmic}
\end{algorithm}
%%%%%%%%%%%%%%%%%%

 \section{
Empirical Understanding of Different Strategies' Impacts on FL's Service Delay
}\label{Sec:SysM1}

In this section, we empirically study and present the impacts of the local computing control and compression techniques (i.e., gradient quantization and weight quantization) on both learning model accuracy and service delay.

%In this section, we present the evaluation of how precision scaling has essential impacts on both learning model accuracy and service delay. 
%\hl{We also illustrate the need for optimized and flexible quantization strategies to accommodate device heterogeneity and minimize service delay. }
%compression control policy

Our experiments use the ResNet20 model on the CIFAR-10 dataset. The computing delay is acquired on two representative mobile devices, Jetson Xavier and Nvidia RTX 8000. Note that the RTX (16.5 TFLOPS) has more computing capabilities than Xavier (1.3 TFLOPS). We set a laptop computer as an FL server and measure the transmission delay of ResNet20 and MobileNetV2 in the Wi-Fi environment. The Wi-Fi uplink transmission rate achieves 102 Mbps on average. More details of the experimental setup can be found in Section \ref{Sec:Result}.  
 
Let $\rho = T^{cp}(32,1)/T^{cm}(32)$ be the ratio of the computing delay per local iteration and the communication delay in full precision. $\rho$ indicates whether the FL task is computing dominant or communication dominant for mobile devices given their computing capabilities and wireless conditions.
 
\subsection{
Impacts of Local Computing Control on the Service Delay
}
\begin{figure}[t]
{\includegraphics[width=1\linewidth]{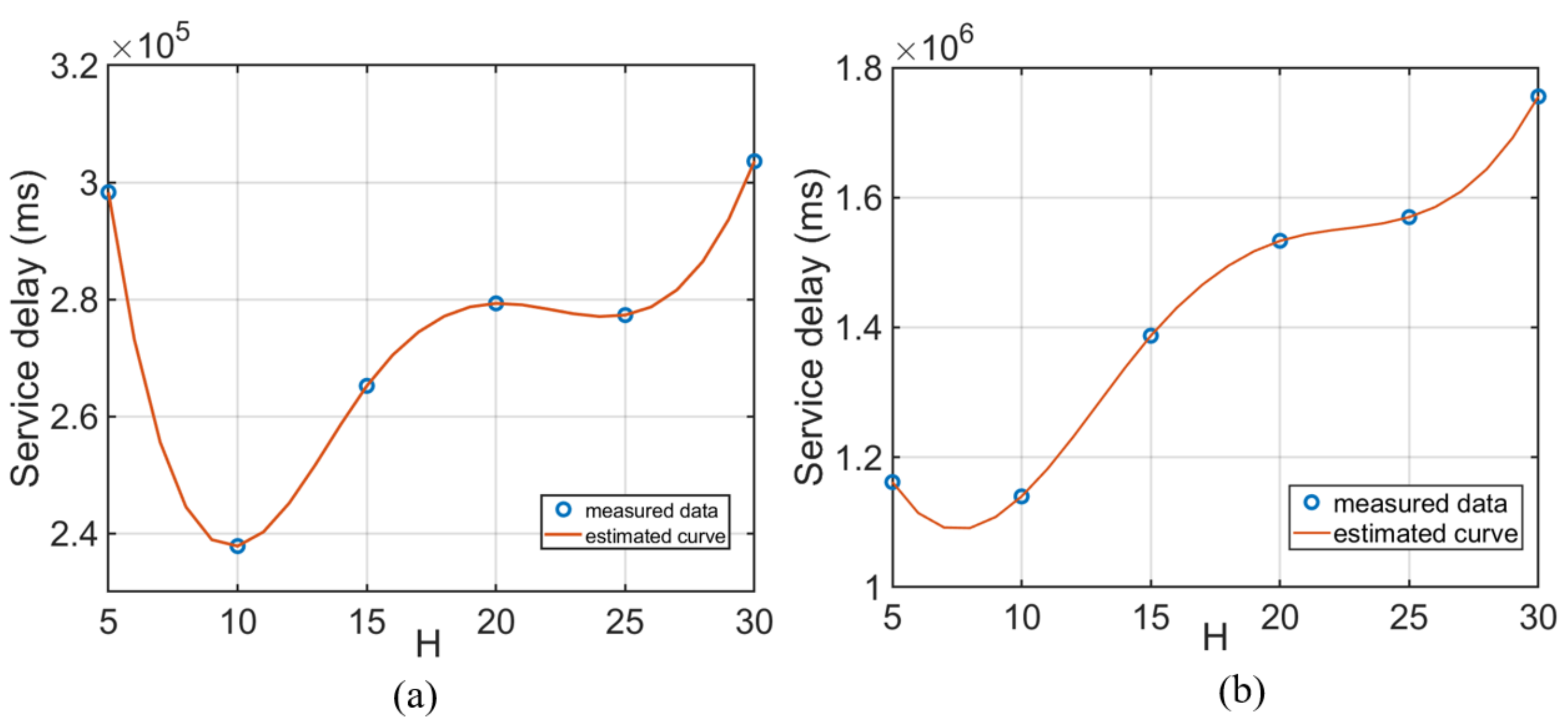}}\vspace{-0.1in}
 \caption{ {The service delay} vs local computing control using ResNet20. The model stops training when the accuracy achieves $98.8 \%$. (a) RTX8000 ($\rho =0.14$); (b) Xavier ($\rho = 0.61$).} \label{fig:Htime}
 %\vspace{-0.2in}
\end{figure}

Local computing control strategies reduce the total number of communication rounds and communication costs by pushing more local computing burdens on the local devices. It has an important influence on model convergence and service delay. Fig. \ref{fig:Htime} depicts the performance (i.e., service delay to reach the target accuracy) on different $H$ and $\rho $ \footnote{Here, we set $q_w$ and $q_g$ as float-point 32 bits (FP32).}.
It is easily observed that selecting a large number of $H$ is not always effective. The service delay first degrades and then quickly increases once $H$ is larger than a certain threshold. With today's wireless transmission rate, the transmission delay of mobile devices is comparable with its computing delay, that is, $\rho = 0.14$ and $\rho = 0.61$ on average for RTX and Xavier, respectively. In this case, increasing $H$ can only reduce the communication time but not necessarily save the service delay. Moreover, the optimal local iteration is related to the ratio $\rho$.  For example, when $\rho$ decreases (e.g., low communication delay or high computing delay), the optimal strategy of minimizing the service delay is to conduct fewer local training iterations and to update the model more frequently. 
 
%%%%figure%%%%

\begin{figure}[t]
{\includegraphics[width=1\linewidth]{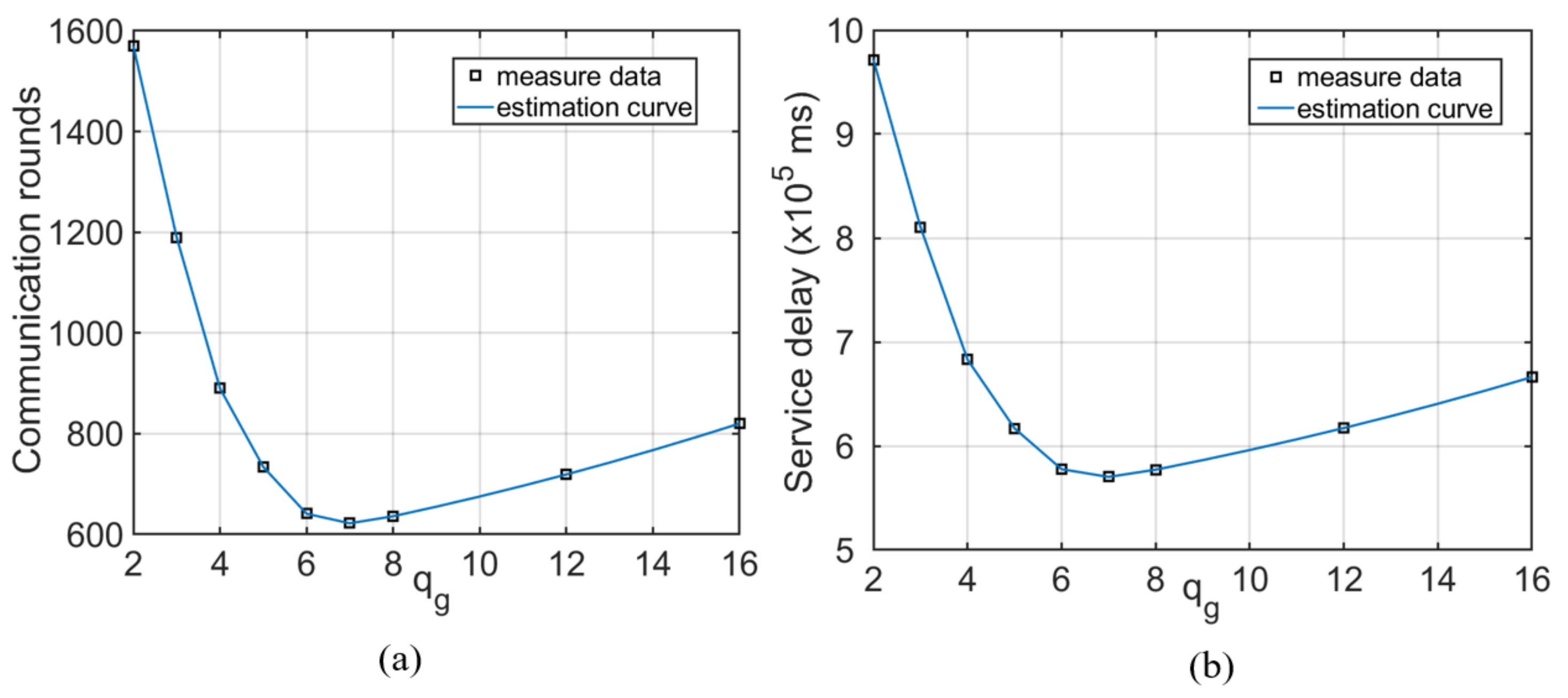}}\vspace{-0.1in}
 \caption{Impact of gradient quantization level using ResNet20 when $H = 10$. The model stops training when the accuracy achieves $98.8 \%$: (a) The number of communication rounds vs gradient quantization level; (b) The service delay vs gradient quantization level.} \label{fig:qtime}
 \vspace{-0.2in} 
\end{figure}
%%%%figure%%%%

\begin{table}[t]
\scriptsize
\centering
  \caption{impact of weight quantization}
  \label{tab:delay}
  \begin{tabular}{|c|c|c|c|l}
  \hline
   $(H, q_g)$ & $(10, 3)$ & $(10, 8)$ & $(10, 16)$ \\
    \hline
    round radio & 0.99 & 0.98 & 0.99  \\
    \hline
    speed up of the service delay & 1.64x & 1.59x & 1.5x \\
    \hline
    $(H, q_g)$ & $(25, 3)$ & $(25, 8)$ & $(25, 16)$\\
    \hline
     round radio & 0.99 & 0.97 & 0.98  \\
    \hline
    speed up of the service delay & 1.68x & 1.65x & 1.61x \\
    \hline
\end{tabular}
\end{table}

\subsection{
Impacts of Gradient Quantization on the Service Delay
}
Similar to the local computing control strategy, gradient quantization reduces the data transmission bits to save communication delay. Unlike the broad discussion on local computing control strategy, the impact of different gradient quantization levels on the service delay is less investigated.

Gradient quantization has a critical influence on both the overall training iterations ($K$) and communication delay per round ($T^{cm}$). For $T^{cm}$, it scales linearly with the quantization level. The extreme case of quantizing data into ternary levels $\{-1,0,1\}$ can ideally reduce 32× communication delay per round than full-precision ones. The impact on $K$ is more complicated than its on $T^{cm}$. We show the impact of the gradient quantization on the communication rounds in Fig. \ref{fig:qtime} (a) and the corresponding service delay in Fig. \ref{fig:qtime} (b). In this subsection, for simplicity, we assign the same gradient quantization level to all the devices. It shows that the extreme reduction in precision (i.e., $q_g$ less than 4 bits) requires more communication rounds to recover the model performance, and thus, the service delay significantly increases. As $q_g$ increases, both the communication rounds and service delay decrease sharply. Surprisingly, we discover that moderate quantization levels (e.g., 5-8 bits) converge faster than the higher-precision counterparts (e.g., 16 bits). The benefit of using moderate quantization levels can be explained by the recent findings on the gradient diversity in distributed learning settings. Gradient diversity measures the difference between the local model gradients and aggregated gradients, and a large gradient diversity encourages active exploration of parameter space~\cite{yin2018gradient}. As such, the noise of moderate quantization properly enlarges the diversity that helps the model converge in lower training loss than the full precision. Hence, by choosing quantization levels judiciously, we can guarantee that model converges fast and accurately.

\subsection{
Impacts of Weight Quantization on the Service Delay
}\label{sec:qw}
Weight quantization reduces data precision to save the data size for memory access of mobile devices. It allows low precision operations to speed up the local computing delay on mobile devices. We investigate the speed-up performance of local training with low precision over the full precision counterparts.\footnote{Jetson Xavier NX features a Volta GPU with 48 Tensor Cores that can accelerate large matrix operations in 16-bits format. Since Tensor Cores only supports FP16 training, we show the speed-up of the service delay when $q_w = 16$.} We perform FL training with different combinations of $(H, q_g, q_w = 16)$ and $(H, q_g, q_w = 32)$. The comparison results are shown in Table \ref{tab:delay}. The “round ratio” denotes the ratio of “Communication rounds for local training with $(H, q_g, q_w = 16)$” and “Communication rounds for local training with $(H, q_g, q_w = 32)$”. This “round ratio” indicates that 16-bit is sufficient precision for most network training. More importantly, the models with the weight quantization of 16 bits can save up to $1.6$x computing delay with almost the same rate as the full precision case. 
 
%, resulting in an 3.5$\times$ faster over FP32 
%Initial experiments reveal that (1) increasing the number of local training iterations does not always reduce the service delay in FL training; and (2) there exists an optimal range of quantization levels to optimize both the model convergence and the overall service delay, which 

\begin{figure}[t] \centering
  %{\includegraphics[width=\linewidth]{System.eps}}
  {\includegraphics[width=0.55\linewidth]{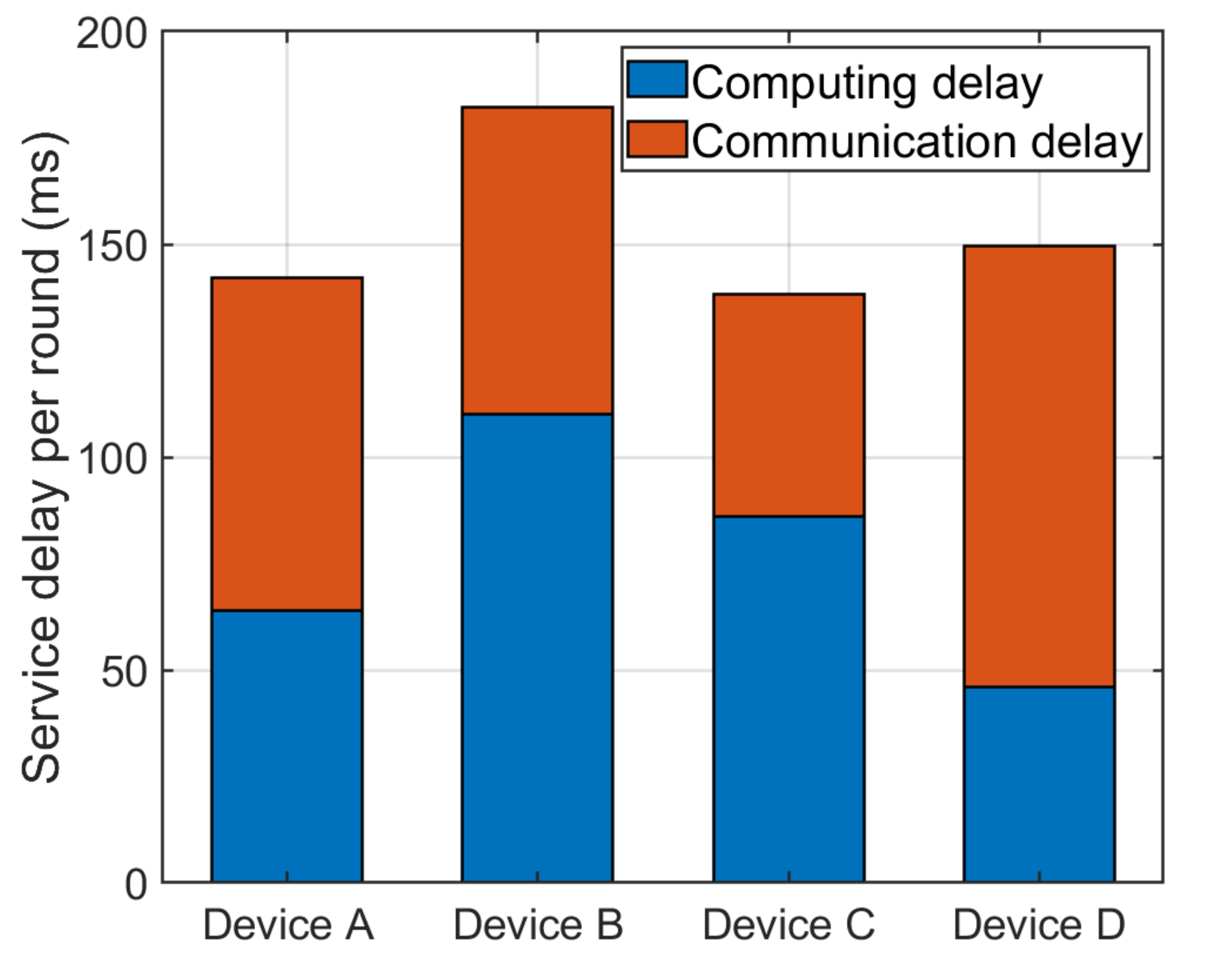}}
 \caption{Service delay vs hetergeneous devices.} \label{fig:Heter}
 \vspace*{-0.2in}
\end{figure}

\subsection{
Impacts of Device Heterogeneity on the Service Delay
}
%We record the local computing delay of Xaviers under different power modes and use iPerf to collect the network throughput. Figure reports that their compute and network capacity are highly different. Further, when we 
We choose different power modes\footnote{For different power modes, the devices operates at different clock frequencies and different activate CPU cores, which would affect computing delay.} of Xaviers to represent the device heterogeneity w.r.t. the local computing capabilities. Fig.~\ref{fig:Heter} reports the service delay of a subset of participating devices when they perform FL training of ResNet20. Hence, we find that: (1) stragglers greatly slow down model updates in practical FL; and (2) optimized quantization strategies should adapt to heterogeneous computing and communication conditions. Here, optimizing the communication delay can greatly benefit Device $D$, but it has limited improvement for the service delay of Device $B$, since it is bottlenecked by the computation delay. 
%\footnote{Device A represents Xavier with working power of 10 W with four cores; Device B of 10 W with two cores; Device C of 15 W with four cores; and Device D of 15 W with six cores.}

In summary, the empirical results above show that the service delay of FL over mobile devices is affected by multiple factors such as the number of local iterations, weight quantization level, gradient quantization level, devices' heterogeneous computing capabilities, different wireless transmission rates of mobile devices for local model updates. In particular, to effectively reduce the service delay, only reducing the computing/communication delay is not good enough, and the trade-off between computing delay and communication delay has to be taken into consideration. Those empirical observations above urge us to jointly consider local computing control, weight quantization strategy, gradient quantization strategy, and device heterogeneity, characterize computing, communications and FL convergence, and develop a optimized solution to minimize the service delay.
 
\section{
Minimizing the Service Delay of FL over mobile devices
}\label{Sec:CovAls}
The experimental observations above reveal that local computing control $H$, gradient quantization levels, $q_g$, and weight quantization levels, $q_w$ play a vital role in the service delay of FL over mobile devices. In this section, we start with discussion on the computing and communication delay model, followed by an approximate analytical relationship between the convergence constraint and the control variables ($q_w$, $q_g$, and $H$). Then, we develop the heterogeneity-aware algorithm to tune $q_w$, $q_g$, and $H$ to balance the trade-off between computing and communication delay of mobile devices in FL training under model convergence.

\subsection{Computation and Transmission Modeling}
\subsubsection{Computation Model}
GPUs are the most commonly used accelerators for DNN computations. We consider the GPUs instead of CPUs in this work for two reasons. First, CPUs cannot support relatively large and complicated model training tasks. Second, GPUs are more efficient than CPUs for on-device DNN training and are increasingly integrated into today's mobile devices. Hence, the computing delay of GPU-based training is comparable to the communication delay of transmitting FL models in high-speed networks (See the example we use in Section III). The local computing of device $n$ involves data fetching from memory and operating arithmetic in core processors. In Section~\ref{sec:qw}, we have observed empirically that weight quantization does speedup the on-device training. Hence, we modify the GPU model in \cite{hong2010integrated} and propose the following delay function to capture the relation among $T^{cp}$, $H$ and $q_w$, 
%{\small }
\begin{equation}
T^{cp}_n (q_{w,n}, H) \triangleq H\left( \alpha_1(q_{w,n})  t_n^{core} + \frac{\alpha_2(q_{w,n})\theta_n^{mem}}{f_n^{mem}} \right) + t_n^0, \label{Tcom}
\end{equation}
where $t_n^{core}$ is the delay coefficient of one training iteration performed in full precision ($q_w=32$), which is determined by the specific DNN structure (e.g., layer configuration and batch size); $\alpha_1(q_{w,n})$ denotes the accelerator factor when the model parameter is represented by $q_{w,n}$ bits. $\theta_n^{mem}$ and $f_{mem}$ denote the number of cycles for device $n$ to fetch data and memory frequency, respectively; $\alpha_2(q_{w,n})$ denotes the scaling factor. $t_n^0$ represents the other component unrelated to training task. 

Here we define function $\alpha_1(q_{w,n})$ in accordance with NVIDIA documentation~\cite{tensore} as
\begin{equation}
    \alpha_1(q_{w,n}) = (1-m) + \frac{m}{(32/q_{w,n})},
\end{equation}
where $m$ denotes the fraction of matrix multiplication and convolution operations that can be accelerated by Tensor Core given a DNN configuration and $(32/q_{w,n})$ represents the ideal speed-up for those operations when quantization level $q_{w,n}$ is selected~\cite{haidar2018harnessing}. Then the speedup of using Tensor Core is  $\frac{1}{\alpha_1(q_{w,n})}$~\cite{tensore}. In term of memory access time, we assume that scaling factor $\alpha_2(q_{w,n})$ are linear functions of data bit-width $q_{w,n}$. This is reasonable since the dataflow (i.e., how data moves in the memory hierarchy) of memory access is fixed given a DNN configuration and hence the number of bits accessed scales linearly with the corresponding bit-widths~\cite{yang2017method}.

%For simplicity, we assume that scaling factor $\alpha_2(q_{w,n})$ are linear functions of data bit-width $q_{w,n}$. This is reasonable since, given a DNN configuration, the dataflow (i.e., how data moves in the memory hierarchy) of memory access is fixed and thus the number of bits accessed scales linearly with the corresponding bit-widths~\cite{yang2017method}.

%we assume that $\alpha_1(q_{w,n})$ and $\alpha_2(q_{w,n})$ are linear functions of data bit-width $q_{w,n}$. This is reasonable since, 

%since the quantization reduces the bit-widths, and the data size scales linearly to the bit representation~\cite{yang2017method}.
%In recent commercial platforms for DNN processing, we can implement the training with mixed precision to speed up the training. 

To validate the results in Eqn. (\ref{Tcom}), we implement two DL models, i.e., ResNet20 and MobileNetv2, on a Xavier NX and RTX with Volta GPU. Table \ref{tab:ctime} shows a comparison between the estimation in Eqn. (\ref{Tcom}) and the actual computing delay measured by Jetson stats~\cite{Jessa}. The result shows that the estimated delay is close to the actual computing delay.

\begin{table}
    \caption{Estimation of one pass computing delay on ResNet20 and MobileNetv2} 
  \label{tab:ctime}
    \centering
    \begin{tabular}{|l|l|l|l|l|}
    \hline
         Model  &\multicolumn{2}{|c|} {ResNet20} &  \multicolumn{2}{c|} {Mobilnetv2} \\ \hline
             & measured & estimated & measured & estimated \\ \hline
        Xavier NX & 74.6 (ms) & 77.3 (ms) & 375 (ms) & 401.3 (ms) \\ \hline
        RTX 8000 & 14 (ms) & 15.2 (ms) & 131 (ms) & 144.6 (ms) \\ \hline
    \end{tabular}
\end{table}
%Xavier & 73 (ms) & 79.8 (ms) & 375 (ms) & 407.9 (ms)
%RTX 8000 & 12 (ms) & 16.8 (ms) & 102 (ms) & 121.5 (ms)
 
\subsubsection{Communication Model}

Let $S$ denote the total number of bits transmitted from devices to the edge server. Given the gradient quantization $q_{g,n}$, we have 
\begin{equation}
    S(q_{g,n}) = s_1 d q_{g,n} + s_0,  
\end{equation}
where $s_0$ and $s_1$ are coefficients determined by additional communication overhead involved in wireless transmission~\cite{khirirat2020communication}. We consider the transmission rate with orthogonal frequency-division multiple access (OFDMA) scheme with total bandwidth $W$. The expected transmission rate\footnote{We do not consider the effect of header information since most deep learning models include a massive number of training parameters (e.g., ResNet has more than $2\times10^7$ bits), indicating that these effects can be ignored.} of mobile device $n$ is given by, 
\begin{equation}
    r_n(\boldsymbol{\lambda}_{n}) =\lambda_{n} W \mathbb{E}_{h_n}\left[\log_2 \left(1+\frac{P_n^{tran} |h_n|^2}{N_0}\right)\right],
\end{equation}
where ${\lambda}_{n}$ represents the resource allocation ratio that satisfied $0 \leq \lambda_{n} \leq 1$ and $\sum_{n=1}^N \lambda_{n} = 1$;  
The expectation is taken over channel fading $h_n$ between mobile device $n$ and the edge server, $N_0$ is white Gaussian noise, and $P_n^{tran}$ denotes the transmitting power. Thus, the communication delay for transmitting the quantized gradients of the quantized model from device $n$ to the edge server is calculated as
\begin{equation}
    T_n^{cm}(q_{g,n}, \boldsymbol{\lambda}_{n}) \triangleq \frac{S(q_{g,n})}{r_n(\boldsymbol{\lambda}_{n})}.\label{Tcomm}
\end{equation} 

\subsection{Relationship Between Convergence and Different Control Variables}\label{Sec:SysM2}
In this subsection, we present the convergence analysis of our proposed FL with local computing control, weight quantization and gradient quantization, and the derived convergence result is used to approximate the effect of different strategies (H, $q_w$, and $q_g$) to the number of training iterations in constraint (\ref{q:convergence}). 

We make the following assumptions on the non-convex function $F_n, \forall n,$ in (\ref{FL_obj}), which are commonly used for non-convex analysis of SGD~\cite{li2019convergence, jiang2018linear, wan2021convergence}. 

\begin{assumption} \label{as:Lsmooth}
  \textit{All the loss functions $F_n, \forall n,$ are differentiable and their gradients are $L$-Lipschitz continuous: for all $x$ and $y \in \mathbb{R}^d$, $\norm{\triangledown F_n(x)- \triangledown F_n(y)}_2 \leq L\norm{x-y}_2, \forall n \in \mathcal{N}$}.
\end{assumption}

\begin{assumption}\label{as:Gdivergence}
  \textit{Assume that $\widetilde{f}_n$ is randomly sampled from device $n$'s local loss functions. The stochastic gradient is unbiased estimator and its variance with a mini-batch of size $M$ :  $\mathbb{E} \norm{ \triangledown \widetilde{f}_n(\boldsymbol{w}^t) - \triangledown F_n(\boldsymbol{w}^t) }_2^2 \leq \frac{\sigma^2}{M}$, and its second moment is $\mathbb{E} \norm{ \triangledown \widetilde{f}_n(\boldsymbol{w}^t)}_2^2 \leq \tau^2, \forall n \in \mathcal{N}$. }
\end{assumption}

\begin{assumption}\label{as:divergence}
  \textit{For heterogeneous data distribution it satisfies:  $\mathbb{E} \norm{ \triangledown {F}_n(\boldsymbol{w}) - \triangledown F(\boldsymbol{w})}_2^2 \leq G^2, \forall n \in \mathcal{N}$.}  
\end{assumption}
 
\begin{theorem}[Convergence of SDEFL] \label{tm:ncvx_convergence}
For SDEFL, under certain assumptions, if the step size satisfies $\eta =\sqrt{\frac{MN}{K}}$, the convergence rate of the proposed scheme satisfies, with the quantization strategies, $q_g, q_w$ and local computing control strategy $H$,
\begin{flalign}
    & \frac{1}{K}  \sum_{k=0}^{K-1} \mathbb{E}  \left[\norm{\triangledown F( {\boldsymbol{w}}^{k})}_2^2\right] \nonumber \\
    &\leq \frac{4\mathbb{E} \left[ F( {\boldsymbol{w}}^{0}) - F( {\boldsymbol{w}}^{K}) \right] }{\sqrt{MNK}} + \frac{ 2 L \sigma^2 (2H\bar{\delta}_g+\bar{p}) }{\sqrt{MNK}}  & \nonumber \\
    & \quad   +\frac{12M L H \bar{\delta}_g G^2}{ \sqrt{MNK} }
    + 2L\sqrt{d}\tau \sum_{n=1}^N p_n^2(\delta_{g,n}+1) \delta_{w,n},\label{eq:cvx_convergence}&
    %
    %&\leq\frac{4\mathbb{E} \left[ F( {\boldsymbol{w}}^{0}) - F( {\boldsymbol{w}}^{K}) \right] }{\sqrt{MNK}} + \frac{ 2 L\sigma^2(H\bar{\delta}_{g} + p) }{\sqrt{MNK}} + 2 L \sqrt{d} \tau \bar{\delta}_{w} & \nonumber      \\  
    %&\quad+\frac{12M L H \bar{\delta}_{g} G^2}{ \sqrt{MNK} } 
     %+ 2 L \sqrt{d} \tau \sum_{n=1}^N p_n^2 \delta_{w,n}\delta_{g,n}, 
     %
    %&  \leq \mathcal{O} (\frac{1+H \bar{\delta}_{g}}{\sqrt{NK}}) +  \mathcal{O}(2 H \sqrt{d} \sum_{n=1}^N p_n^2 \delta_{w,n}\delta_{g,n}),& \\
    %  &  \leq   \frac{A_1 + A_0 H \sum_{n=1}^N p_n^2\delta_{g,n}}{\sqrt{NK}} +  B_0 \sum_{n=1}^N p_n^2 \delta_{w,n}\delta_{g,n},&
\end{flalign}
where $\bar{\delta}_{g} = \sum_{n=1}^N p_n^2\delta_{g,n}$, $\bar{p} = \sum_{n=1}^N p_n^2$ and $\bar{\delta}_{w} = \sum_{n=1}^N p_n^2\delta_{w,n}$. 
\begin{proof}
See Appendix~\ref{proof:tm}.
%please refer to the detailed proof\footnote{https://github.com/qweriuo/proof/blob/main/proof.pdf} at the Github. 
\end{proof}
\end{theorem}
Theorem~\ref{tm:ncvx_convergence} implies that the convergence rate of SDEFL is $\mathcal{O}((1+\bar{\delta}_{g}H)/\sqrt{NK})$. It converges at the same rate as the full-precision SGD, but only to a non-zero error (the last term) associated with the quantization resolution $\delta_g, \delta_w$, and model dimension $d$. Hence, weight quantization makes FL converge to a neighborhood of the optimal solution without affecting the convergence rate. The 16-bit weight quantization does not impede down convergence performance, which is consistent with our empirical observations of weight quantization. Furthermore, 16-bit weight quantization has little effect on model learning performance. While gradient quantization and local iterations would affect the convergence by a factor of $\mathcal{O}( H \bar{\delta}_g)$. Given a target model accuracy, a higher value $H$ and a lower value $q_{g,n}$ result in a higher bound of communication rounds.

\subsection{Problem Formulation}
Given the computing model and the communication model above, we reformulate the problem in (\ref{eq:time_obj})-(\ref{eq:convergence_ori}) as follows
\begin{flalign}
     &\min_{\substack{H, K, \\ q_{g,n}, q_{w,n}}} \quad  T_{tot}      \label{Objective}  \\
     &\text{s.t.,} \quad 
       \frac{A_1+A_0H\sum_{n=1}^N p_n^2\delta_{g,n}}{\sqrt{NK}} + C_0 \sum_{n=1}^N p_n^2 \delta_{w,n} \nonumber\\
       &\qquad\quad + B_0 H \sum_{n=1}^N p_n^2\delta_{g,n}\delta_{w,n}  \leq \epsilon, \label{q:convergence} \\
     & \quad\quad H \in \mathcal{H}, q_{g,n} \in \mathcal{Q}_g, q_{w,n} \in \mathcal{Q}_w, \forall n \in \mathcal{N}. \label{RQ1}
\end{flalign}
Here, we leverage the convergence upper bound as an approximation of Eqn.~(\ref{eq:convergence_ori}) to achieve $\epsilon$-global model convergence. Constraint (\ref{q:convergence}) represents that after $K$ training iterations, the training loss should be smaller than a pre-set threshold $\epsilon$. For simplicity, we use the coefficients $A_1$, $A_0$, $B_0$, and $C_0$ to characterize the loss function related properties and statistical heterogeneity of non-i.i.d. data in Theorem 1. The coefficients can be estimated by using a sampling set of experimental training results. Constraints (\ref{RQ1}) indicate that optimization variables take the values from a set of non-negative integers. We relax the $K$, $q_w$, and $q_g$  as continuous variables for theoretical analysis, which later are rounded back to the nearest integer. For the relaxed problem, we observe that the variable $K$ is monotonously decreased with the objective function. Hence, for optimal $K$, constraint (\ref{q:convergence}) is always satisfied with equality, and we can obtain $K$ as
\begin{equation}
  K = \frac{ (A_1+A_0H\sum_{n=1}^N p_n^2\delta_{g,n})^2}{N(\epsilon-B_0 H \sum_{n=1}^N p_n^2\delta_{g,n}\delta_{w,n} - C_0\sum_{n=1}^N p_n^2 \delta_{w,n})^2}. \label{round}
\end{equation}

For notation simplicity, we further simplify the description of GPU time model as $T_n^{cp} = H(\beta_n^1 q_{w,n} +\beta_n^0)$, where $\beta_n^0 = t_n^0 + (1-m) t_n^{core}$ and $\beta_n^1 = \theta_n^{mem}/f_n^{mem}+ m t_n^{core}/32$ and communication model as $u_{n}^1 = {d s_1}/r_n$ and $u_{n}^2 ={s_0 }/r_n$. By substituting the (\ref{round}) into its expression, we reformulate the problem (\ref{Objective})-(\ref{RQ1}) as follows,
{%\small
\begin{flalign}
     \min_{\substack{H, q_{g,n},q_{w,n}}}  & \quad \frac{ (A_1+A_0H\sum_{n=1}^N p_n^2\delta_{g,n})^2}{HN(\epsilon- \sum_{n=1}^N p_n^2\delta_{w,n} (B_0 H\delta_{g,n} +  C_0)^2} \nonumber \\        
    &\quad \cdot \max_{n \in \mathcal{N}}  \{u_{n}^1 q_{g,n} + u_{n}^2 + H(\beta_n^1 q_{w,n} +\beta_n^0) \} \label{reObj}\\
     \text{s.t.}  \quad & H \in  \widetilde{\mathcal{H}}, q_{g,n} \in \widetilde{\mathcal{Q}}_g, q_{w,n} \in {\mathcal{Q}}_w, \forall n \in \mathcal{N}, \label{q:rq1}
\end{flalign} 
}
where $\widetilde{\mathcal{H}}$ and $\widetilde{\mathcal{Q}}_g$ are the relaxed set. The problem (\ref{reObj})-(\ref{q:rq1}) is non-convex problem. The non-convexity arises from (1) the min-max formulation; and (2) multiplicative form of $q_{w,n}$, $q_{g,n}, \forall n \in \mathcal{N}$ and $H$ in both the objective function and constraints, which makes the optimization NP-hard to solve. 

To reduce the computational complexity, we convert the optimization problem in (\ref{reObj})-(\ref{q:rq1}) to a geometric programming problem, which is then converted into a standard convex problem via log-sum-exp forms~\cite{chiang2005geometric}. 
%First, based on the empirical observation of the impact of weight quantization (see Section~\ref{sec:qw}), mixed-precision training not only maintains similar training iterations and model accuracy as the full precision training but also reduces service delay. Hence we consider those devices with the capability of mixed-precision training will determine $q_{w,n} =16$ as their desired weight quantization strategy. In the following analysis, the $q_{w,n}$ is not considered as an optimization variable. 
Then we introduce a slack variable $\Psi$ such that %$\Psi =T(\boldsymbol{q}_{g}, H)$ and hence $T_n \leq \Psi, \forall n$.
{\small
\begin{flalign}
\min_{\substack{\Psi, \phi, H,\\ q_{g,n}, q_{w,n}}} & \quad \Psi, \label{reObj1}  \\
\text{s.t.}\quad  %&\quad (\ref{q:rq1}), \nonumber \\    
& \frac{\phi^2}{HN} \left(u_{n}^1 q_{g,n} + u_{n}^2 + H(\beta_n^1 q_{w,n} +\beta_n^0) \right) \leq \Psi,  \forall n \in \mathcal{N}, \label{ieq:t_sqrt}\\
& 0 \leq \frac{A_1+A_0H\sum_{n=1}^N p_n^2\delta_{g,n}}{\epsilon- \sum_{n=1}^N p_n^2\delta_{w,n} (B_0 H\delta_{g,n} +  C_0)} \leq \phi,
\end{flalign} }
According to the Definition~\ref{def:q}, we can assume that $\delta_{g,n} $ and $\delta_{w,n}$ decrease monotonically with $q_{g,n}$ and $q_{w,n}, \forall n$, respectively. We add another constraints\footnote{This constraint can be easily replaced with coefficient functions of different quantization schemes.}, for $\xi \in \{g, w\}$, 
\begin{flalign}
    \delta_{\xi,n} &= ({1+\sqrt{2d-1}})/(2^{q_{\xi,n}}-1 ), n \in \mathcal{N}.
\end{flalign}
We then introduce additional auxiliary variables $v_{\xi,n}$, where $q_{\xi,n} = \log_2(1 + 1/v_{\xi,n}) \approx log(2)/v_{\xi,n}$ and $\delta_{\xi,n} = v_{\xi,n}({1+\sqrt{2d-1}})$. Further, since $\epsilon$ is a constant and we know that $\epsilon \geq B_0H\sum_{n=1}^N p_n^2\delta_{g,n}\delta_{w,n}$, then constraint (\ref{ieq:t_sqrt}) is equivalent to
{\small
\begin{flalign}
& \frac{A_1+A_0'H\sum_{n=1}^N p_n^2v_{g,n}}{\phi\epsilon} + \frac{\sum_{n=1}^N p_n^2 v_{w,n} (B_0'Hv_{g,n}+C_0') }{\epsilon} \leq 1,
%& \frac{A_1+A_0'H\sum_{n=1}^N p_n^2v_{g,n}}{\phi\epsilon} + \frac{B_0'H\sum_{n=1}^N p_n^2 \delta_{w,n}}{\epsilon} \leq 1,
\label{reieq:t_sqrt}
\end{flalign}} 
where {\small$A_0' = A_0({1+\sqrt{2d-1}})$}, {\small$B_0' = B0({1+\sqrt{2d-1}})^2$}, and  {\small$C_0' = C_0({1+\sqrt{2d-1}})$}. With auxiliary variables $v_{g,n}$ and $v_{w,n}, \forall n \in \mathcal{N}$, we can observe that problem (\ref{reObj1})-(\ref{reieq:t_sqrt}) satisfies geometric programming (GP) and transform into convex functions. Hence, it can be efficiently solved optimally by the edge server through the interior point primal dual method~\cite{nesterov1994interior}.

%, without the loss of generality

The edge server is in charge of solving the optimization in (\ref{reObj1})-(\ref{reieq:t_sqrt}). It is practical because in the FL protocol in \cite{bonawitz2019towards} requires mobile devices to check in with the FL server first. Hence, the FL server can collect the information from mobile devices and find the optimal strategy ($H$, $q_{w,n}$ and $q_{g,n}$) for each device via an optimization algorithm.

\begin{figure}[t]
{\includegraphics[width=1.0\linewidth]{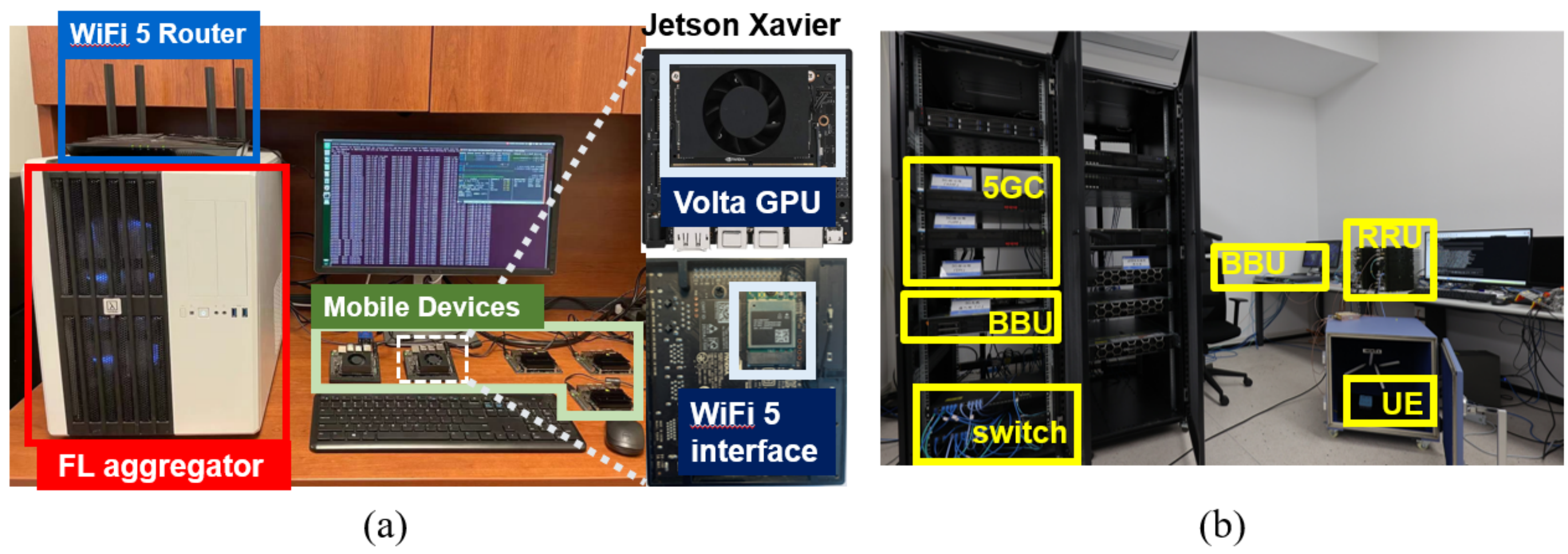}}
 \caption{The SDEFL system configuration: (a) The SDEFL testbed with RTX 8000 (FL aggregator) and Jetson Xavier kits (FL clients) in our lab; (b) The 5G emulation platform in the Purple Mountain Laboratories, Nanjing, China.} \label{fig:testbed}
 \vspace{-0.15in}
\end{figure}
% in the Purple Mountain Laboratories, Nanjing, China.

\section{SDEFL Performance Evaluation}\label{Sec:Result}
In this section, we illustrate the setup of testbeds, and conduct extensive experiments/simulations using different learning models and wireless transmission techniques (i.e., Wi-Fi 5 and 5G) to evaluate the performance of the proposed SDEFL in terms of service delay and learning accuracy.

\begin{figure*}[t] \centering 
  \subfigure[Service delay vs $H$. \label{fig:TvQ}]
  {\includegraphics[width=.241\textwidth]{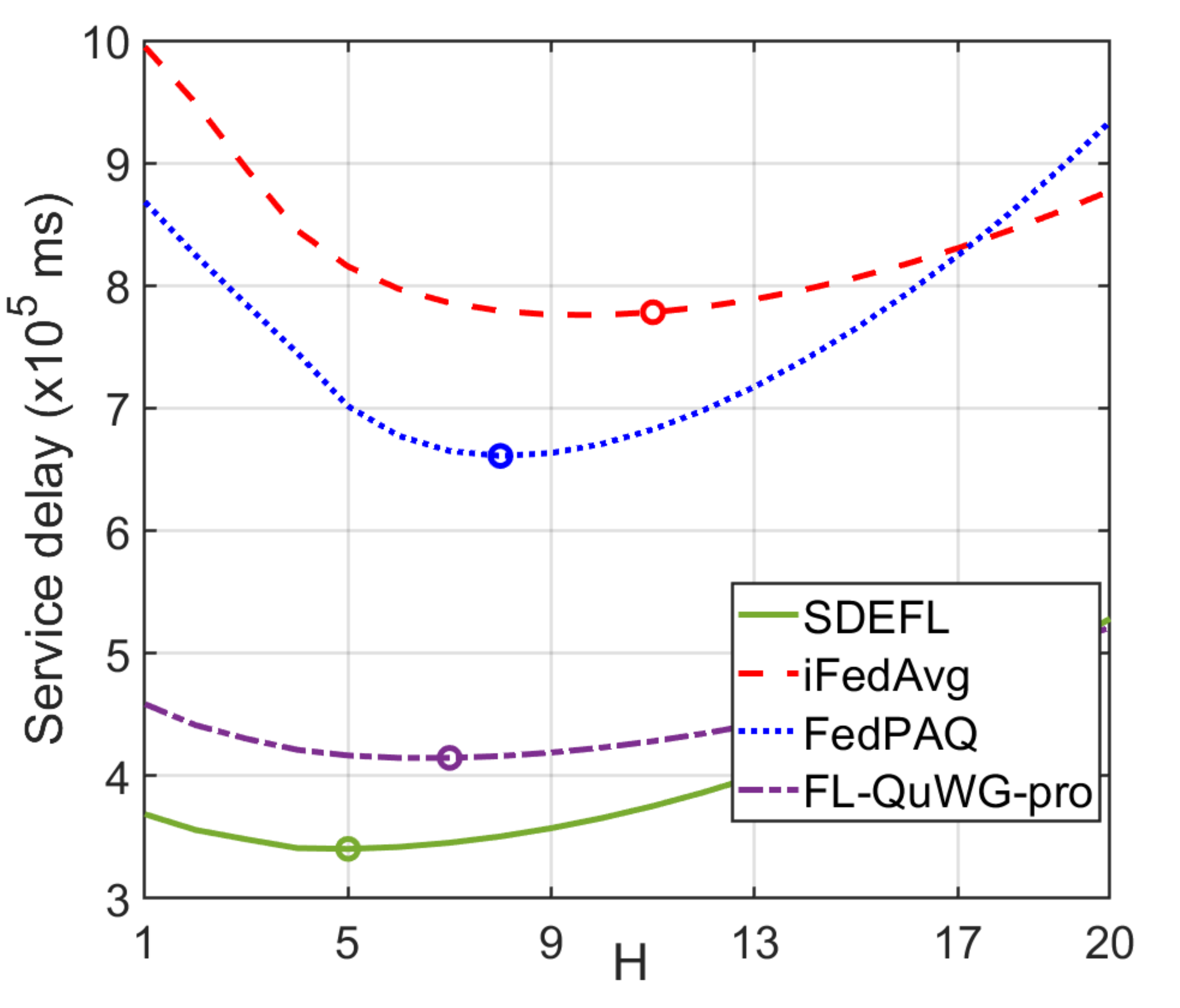}}
  \subfigure[Testing accuracy vs service delay. \label{fig:TvACC}]
  {\includegraphics[width=.241\textwidth]{ 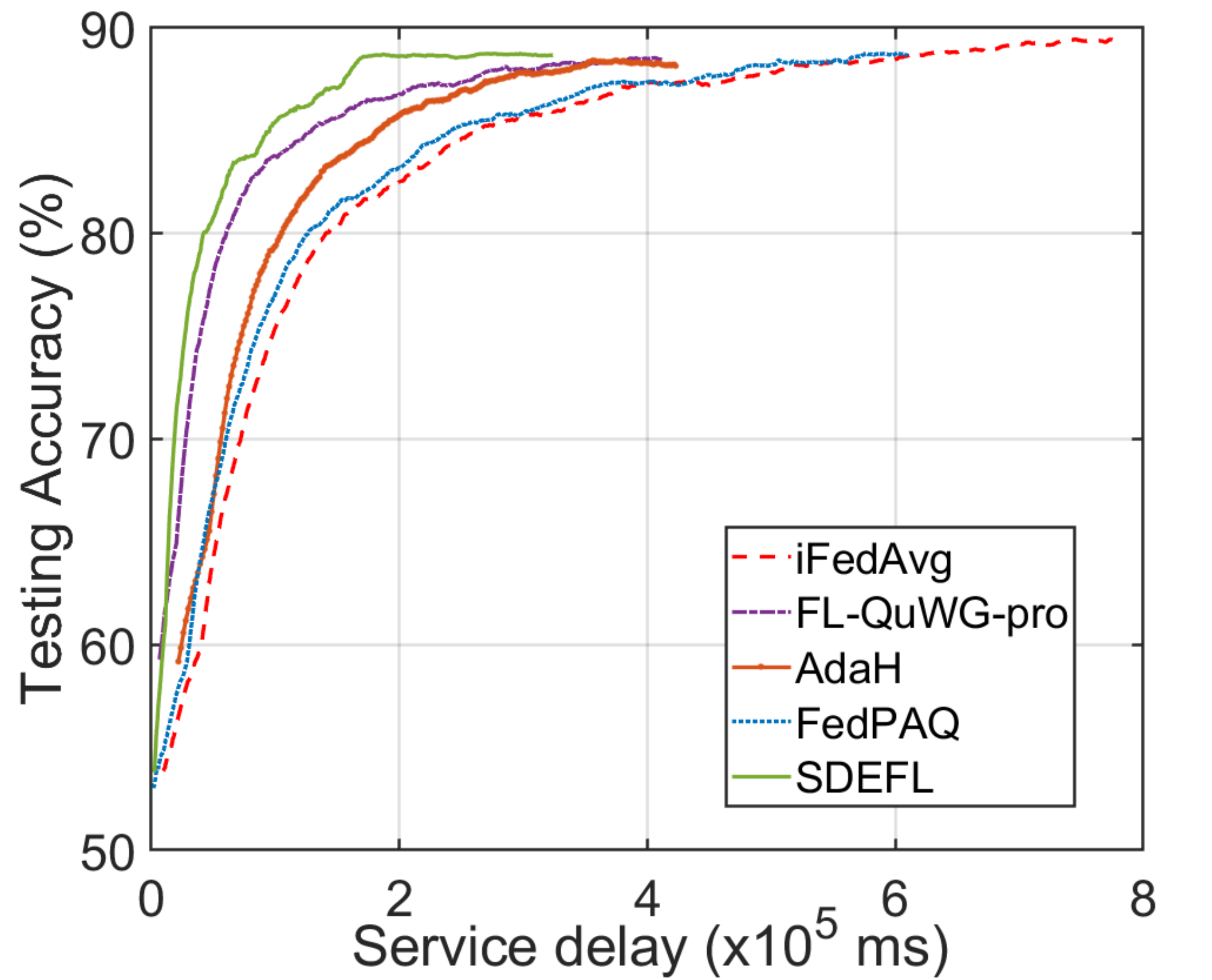}}
  \subfigure[Service delay vs $H$. \label{fig:TvQ1}]
  {\includegraphics[width=.241\textwidth]{ 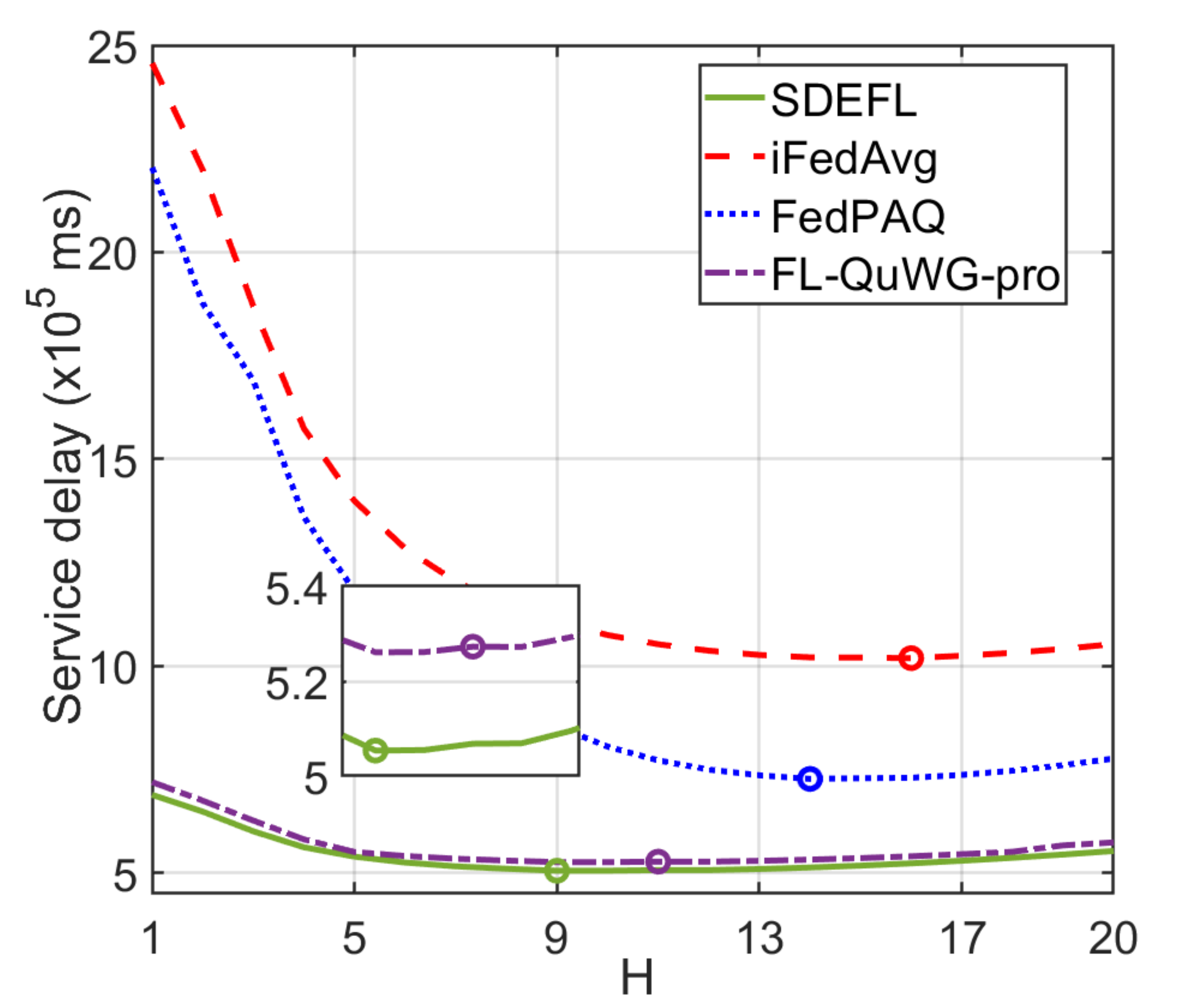}}
  \subfigure[Testing accuracy vs service delay. \label{fig:TvACC1}]
  {\includegraphics[width=.241\textwidth]{ 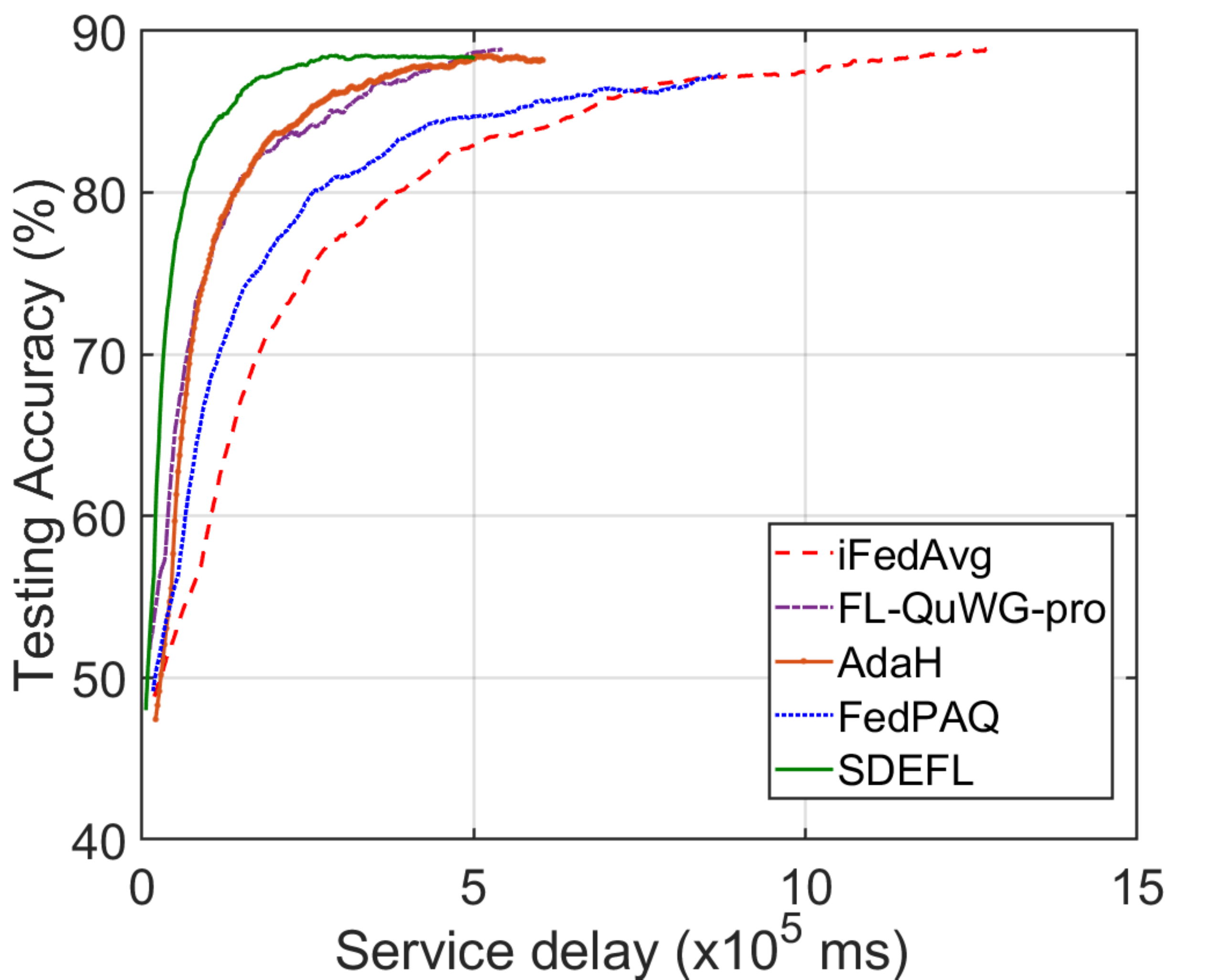}}
  \vspace{-0.1in}
  \caption{Training performance of ResNet20. The estimation of $A_0 = 0.35, A_1 =32.3$, $B_0 = 0.001$, and $C_0 = 0.06$. (a)-(b): Experimental Results (Wi-Fi 5: 88 Mbps); (c)-(d): Emulation Results (5G: 14 Mbps). The dots show that the optimal $H$ in different schemes.}\label{Fig:sim}
  \vspace{-0.2in}
\end{figure*}

\subsection{Experimental Setup}

\subsubsection{System Configurations}
We evaluate the SDEFL's performance in both testbeds (for mobile devices with Wi-Fi 5 transmission capability) and a simulated environment (for mobile devices with 5G transmission capability). Our SDEFL testbed consists of an NVIDIA RTX 8000 as the edge server and several Jeston Xavier kits as mobile devices, as shown in Fig.~\ref{fig:testbed}(a). Each Xavier has a Volta architecture, with 384 NVIDIA CUDA cores and 48 Tensor cores, and 8GB GPU memory. All devices are wirelessly connected via Wi-Fi 5 according to the WebSocket~\cite{websoc} communication protocol. We also adopt the iPerf3 network speed test tool to measure the achievable bandwidth of the wireless transmissions. As for the simulated environment, since Jeston Xavier doesn't support 5G transmissions\footnote{The NVIDIA Jetson Xavier NX developer kits we have in the lab don't support Wi-Fi 6 or 5G transmission.}, we have to separately test the computing delay and 5G transmission delay of SDEFL. Here, we measure the computing delay using Xaviers in our lab, measure the transmission delay using a 5G emulation platform, and combine the results to evaluate the service delay of SDEFL. The system configuration is shown in Fig.~\ref{fig:testbed}.
 % from the Purple Mountain Laboratories, Nanjing, China
 %\hl{We choose Jetson Xavier NX as mobile device since Xavier NX devices equipped with NVIDIA Tensor cores are the only commercial mobile GPUs that provide actual speedup with training with FP16 as far as we acknowledge.}

\subsubsection{Learning Datasets and Models}
We evaluate our results on the CIFAR-10 dataset, which consists of 50000 training images and 10000 test images in 10 classes. Without loss of generality, we choose to train deep neural networks ResNet20 and MobileNetv2 from the scratch, since these two DNN models have different architectures and numbers of model parameters. For all experiments, we set the batch size $M = 128$. In each round of FL, we consider 10 participating mobile devices, which run $H$ steps of SGD in parallel. We employ an initial learning rate $\eta = 0.1$ with a fixed decay rate of 0.996. Each result is averaged over 5 experiments.

\subsubsection{Implementation}

We implement SEDFL and the other baselines by building on top of Flower~\cite{beutel2021flower}. In total, we add 400 lines to perform weight and gradient quantization. In particular, we use NVIDIA’s APEX library for quantization implementation. The optimal local computing control, weight quantization, and gradient quantization strategies are obtained by solving the SDEFL optimization problem in (\ref{reObj1})-(\ref{reieq:t_sqrt}). The SDEFL optimization is conducted using ``fmincom-geometric" solver. It takes 30 ms to solve the proposed optimization by using MATLAB, which is definitely affordable for the FL edge server. 
%Due to the fact that the Tensor cores in Xaiver NX only support hardware acceleration for training with FP16, we will not consider any weight quantization level smaller than 16-bit in the test-bed experiments. Moreover, in Section \ref{Sec:SysM1}, we show that using 16-bit weight quantization can fully restore the model performance (i.e., model accuracy and the required training iterations) of the full precision counterparts. Hence those devices with the capability of mixed-precision training will determine $q_{w,n} =16$ as their preferred weight quantization strategy in the test-bed experiments. 
Besides quantization implementation, there are some other implementation details as follows.

%, and corresponds to 0.8\% of the average time for aggregation rounds
%As current Tensor cores on Volta GPU only supports hardware speedup for mixed precision training with FP16, we do not consider any weight quantization level lower than 16-bit in the following experiments due to hardware limitation.

\noindent \underline{\emph{Experiments with i.i.d data}}: We set up the testbed and conduct the first series of experiments using ResNet20 and MobilenetV2 with CIFAR10 dataset with $N = 10$ participating devices with homogeneous data distribution. In particular, we uniformly distribute 6,000 data samples and 10 classes among the participating devices. %For the SDEFL testbed in our lab, the achievable Wi-Fi 5 transmission rate is 88 Mbps on average.

\noindent \underline{\emph{Experiments with non-i.i.d data}}:  With the same testbed, we conduct the second series of experiments using ResNet20 and CIFAR10 dataset with heterogeneous data distribution. To generate unbalanced data, we sample the number of data samples via a lognormal distribution with a standard deviation from $\{0.3, 0.6\}$, and each device contains only 4 digits labels.

\noindent \underline{\emph{Implementation of wireless transmissions}}: Using the 5G emulation platform, we measure the achievable 5G uplink transmission rate, which is 14 Mbps on average with standard deviation (SD) 3.85 Mbps. Using the SDEFL testbed in our lab, we measure Wi-Fi 5's achievable uploading transmission rate from Xaviers to the RTX server, which is 88 Mbps on average with SD 16 Mbps.

% in the Purple Mountain Laboratories

\subsubsection{Peer Schemes for Comparison}
We compare our proposed {SDEFL} scheme with the following FL schemes:
(1)\textit{iFedAvg}: Similar to the work in \cite{luo2021cost}\footnote{In this work, Luo et al consider the client selection and the optimal number of participant is full participants when we only consider the service delay. Hence, we only consider the optimal value of $H$.}, all the devices follows FedAvg algorithm with the optimal value of $H$ and do not consider weight and gradient quantization. 
(2) \textit{FedPAQ}: In FedPAQ \cite{reisizadeh2020fedpaq}, all the devices perform local training with full precision weight and transmit the quantized version of model updates to the edge server. In this scheme, all clients are assign with the same gradient quantization level. (3) \textit{AdaH}: Similar to the work in \cite{wang2019adaptive}, {all the devices follow the same dynamic $H$ policy. The dynamic $H$ policy gradually decreases $H$-value as $H^t = \sqrt{\frac{F^t}{F^0}} H^0$, and we set $H^0$ = $30$.} (4) \textit{FL-QuWG-pro}: We assign different gradient quantization levels proportionally according to the allocated bandwidth of each mobile device. 
% The optimal strategies are optimized by solving a simplified version of the problem (\ref{reObj}). 

\subsection{Results and Analysis}
%\textcolor{red}{Any comparison with resource allocation only algorithm such as Ref[10]? Besides, some reference typos, e.g., Ref[7].}

\subsubsection{Service Delay and Learning Accuracy} 
Fig.~\ref{Fig:sim} shows the service delay for reaching the target loss of 0.15 under different numbers of local iterations $H$ using our testbed with Wi-Fi 5 transmissions. In Fig.~\ref{Fig:sim}(a) and Fig.~\ref{Fig:sim}(c), where the communication and computing delay is comparable, the optimal $H$ exists (not extremely big/small), which further confirms our empirical and theoretical analysis. Besides, from Fig.~\ref{Fig:sim}(b) and Fig.~\ref{Fig:sim}(d), we observe that with properly selected gradient quantization strategy, \textit{FedPAQ}, \textit{FL-QuWG-pro}, and SDEFL can effectively reduce the service delay. Moreover, our proposed SDEFL scheme outperforms the others: 1) It is because SDEFL is more flexible than \textit{FedPAQ} in terms of quantization decision making, which can better fit the heterogeneous communication conditions across participants. 2) SDEFL outperforms \textit{AdaH} since the dynamic policy in \textit{AdaH} only depends on the training loss and does not consider the trade-off between communication and computing delay. It also ignores the heterogeneous resource across participants. Different from it, the proposed SDEFL focuses on the trade-off between communication and computing delay and device heterogeneity to further reduce the service delay.
3) Compared with \textit{FL-QuWG-pro}, it demonstrates the effectiveness of the proposed optimization that can find the optimal strategies for different mobile devices. It shows when reaching FL convergence, SDEFL can reduce  56\% service delay with 0.3\% accuracy loss compared with \textit{FedAvg}, and reduce round 50\% service delay with 0.23\% accuracy loss compared with \textit{FedPAQ}. 
%They even converge slightly faster than the full precision counterparts. 
%Hence the service delay of \textit{AdaH} is restricted by the system stragglers.

\subsubsection{Impacts of Wireless Transmission Rates} 
As the measured transmission rate of 5G is slower than that of Wi-Fi 5, the SDEFL with 5G transmissions has larger service delay as shown in Fig.~\ref{Fig:sim}. Actually, different transmission rates yield different workload allocations between ``working" and ``talking" to achieve the minimum service delay, which is reflected by the computing delay/communication delay ratio $\rho$ defined in Sec.~\ref{Sec:SysM1}. Here, $\rho$ equals to 0.9 for SDEFL with Wi-Fi 5 transmission, and $\rho$ equals to 0.1 for SDEFL with 5G transmission. In Fig. \ref{Fig:sim} (a) and (c), we find that the optimal value of $H$ decreases as $\rho$ increases (from 0.1 to 0.9). The reason is that SDEFL prefers to reduce the service delay by ``talking" more in the case of the high wireless transmission rate, i.e., the accumulated wireless transmission delay is comparable or smaller than the accumulated local computing delay for mobile devices during FL training. It further verify our observation and theoretical analysis about the trade-off between communication and computation in FL over mobile devices. Besides, for different $\rho$ values, the proposed SDEFL scheme outperforms other schemes. 
%achieves near-optimal performance compared to the empirical optimal solutions and 

\begin{table}[t]\centering \scriptsize
\caption{Testing accuracy with different FL schemes.   \label{tab:acc}}
\begin{tabular}{|c|c|c|c|c|}
\hline
Model& Methods  & Acc.(\%) & \makecell[c]{Service delay \\ in (\ref{reObj}) (ms)} & \makecell[c]{Measured \\ delay (ms)} \\ \hline
\multirow{5}{*}{ResNet20}    & iFedAvg~\cite{luo2021cost}   & \bf{89.98} & 776435     & 714653 \\   \cline{2-5} 
                             & FedPAQ~\cite{reisizadeh2020fedpaq}   & 87.15      & 682118     & 655170 \\  \cline{2-5}
                             & FL-QuWG-pro   & 87.15      & 415864     & 362726 \\  \cline{2-5}
                             & SDEFL    & 88.64      &\bf{340273
} & \bf{316819} \\ \hline
\multirow{5}{*}{ {MobileNetv2}} & iFedAvg~\cite{luo2021cost}   & \bf{88.19} & 1933978    & 1765978 \\  \cline{2-5} 
                             & FedPAQ~\cite{reisizadeh2020fedpaq}   & 82.62      & 1808992    & 1704431 \\  \cline{2-5}
                             & FL-QuWG-pro   & 87.15      & 1109046    & 1011043 \\  \cline{2-5}
                             & SDEFL    & 83.47      &\bf{931106} & \bf{871240} \\ \hline
\end{tabular}
\vspace{-0.2in}
\end{table}

%Relative service delay with comp. heterogeneity.
%Service delay with comm. heterogeneity. 241
\begin{figure*}[t] \centering 
  \subfigure[Service delay with DH. \label{fig:non-iid}]
  {\includegraphics[width=.241\textwidth ]{ 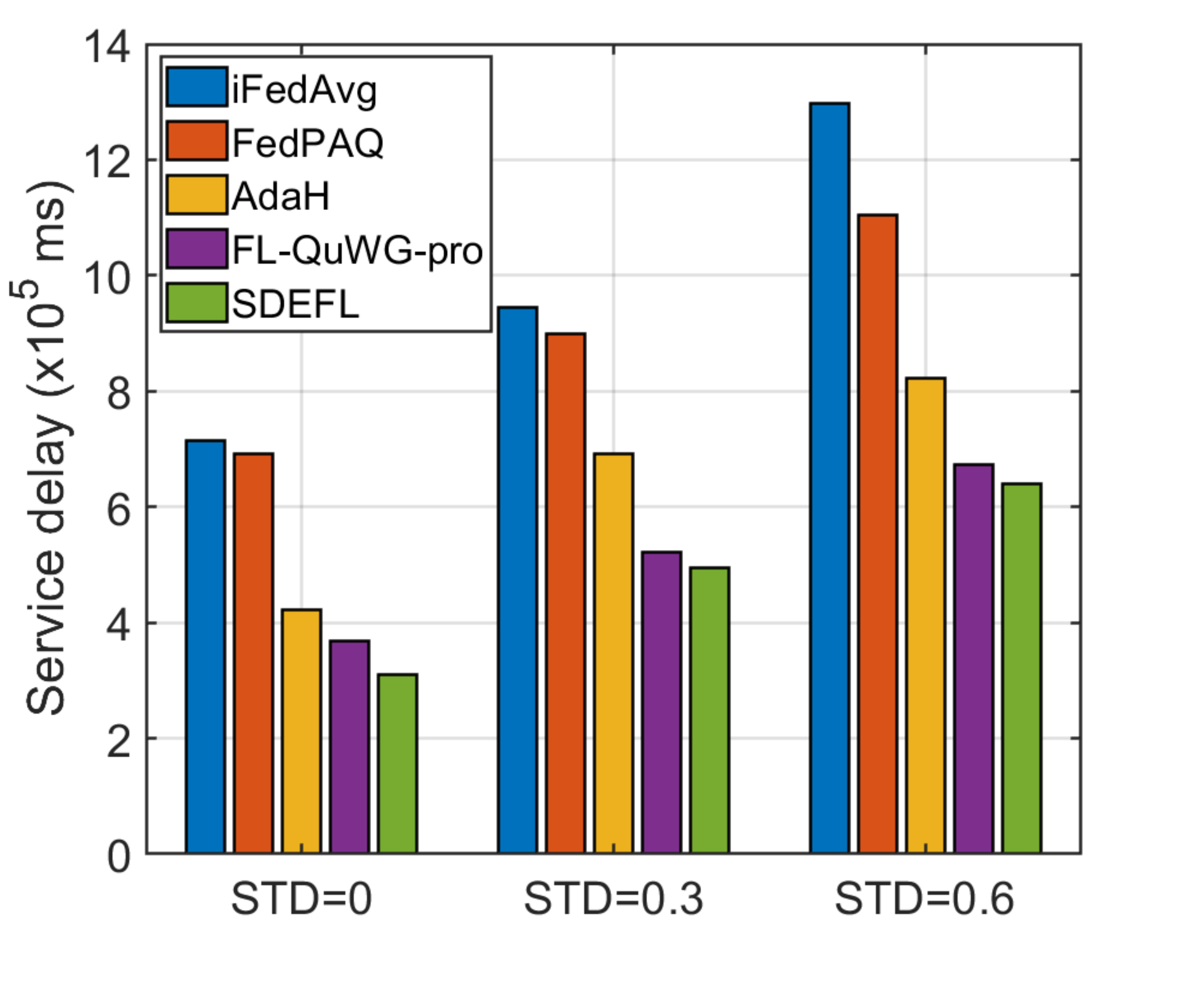}}
  \subfigure[Service delay with $\gamma_{cp}$.\label{fig:comp-heter1}] 
  {\includegraphics[width=.241\textwidth]{ 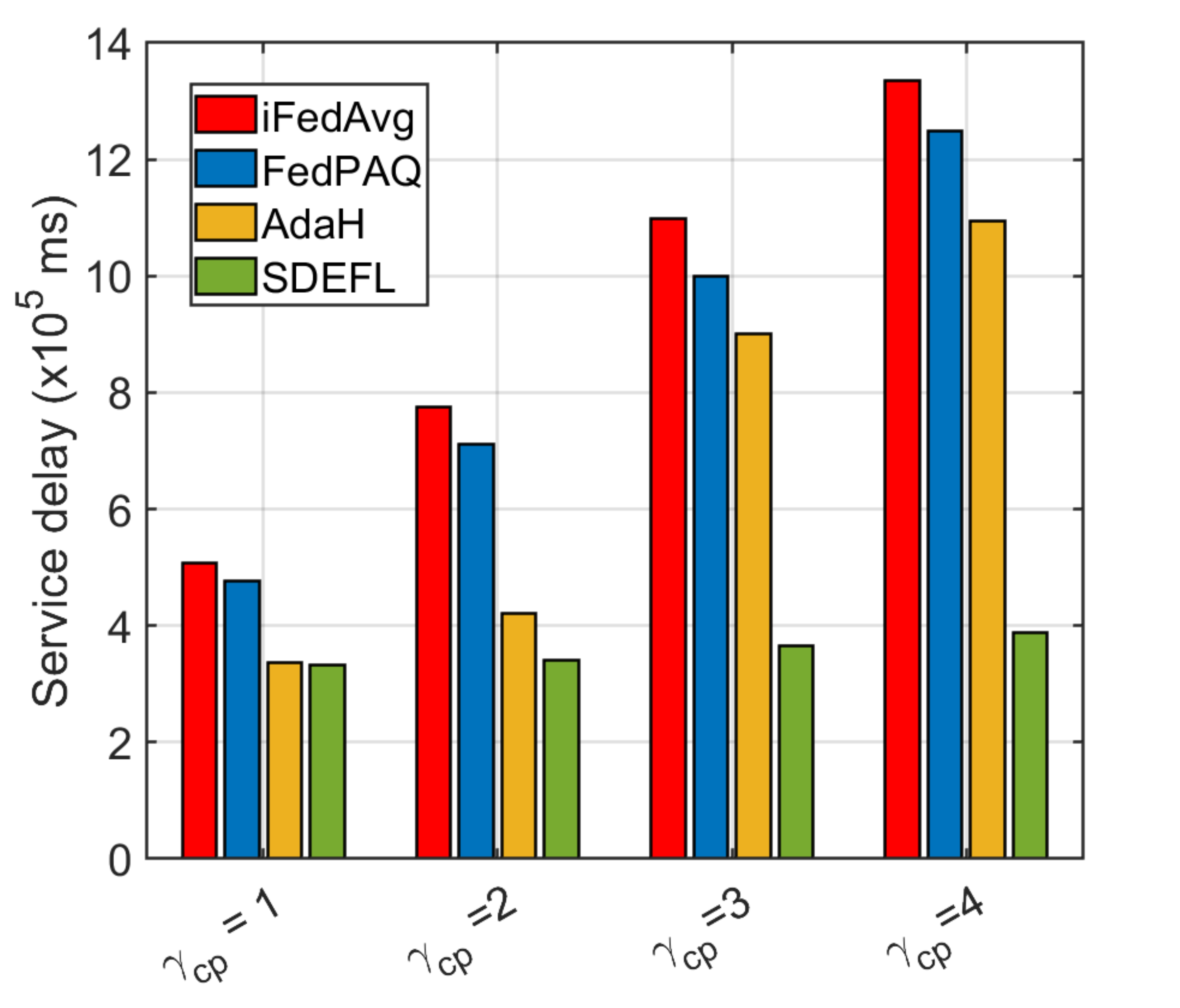}}
  \subfigure[Relative service delay w.r.t $\gamma_{cp}$ \label{fig:comp-heter}]
  {\includegraphics[width=.241\textwidth]{ 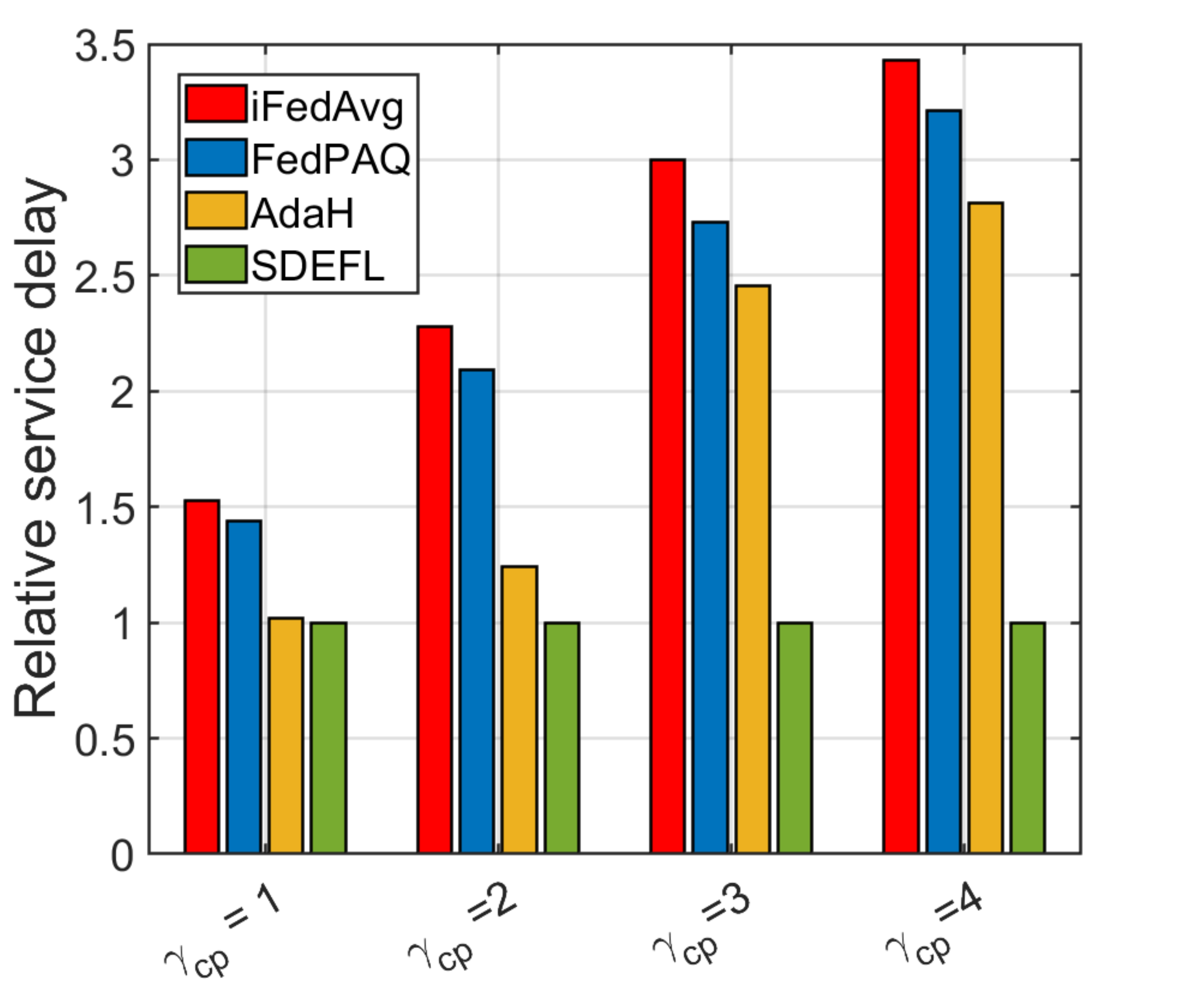}}
  \subfigure[Relative service delay w.r.t $\gamma_{cm}$ \label{fig:comm-heter}]
  {\includegraphics[width=.241\textwidth]{ 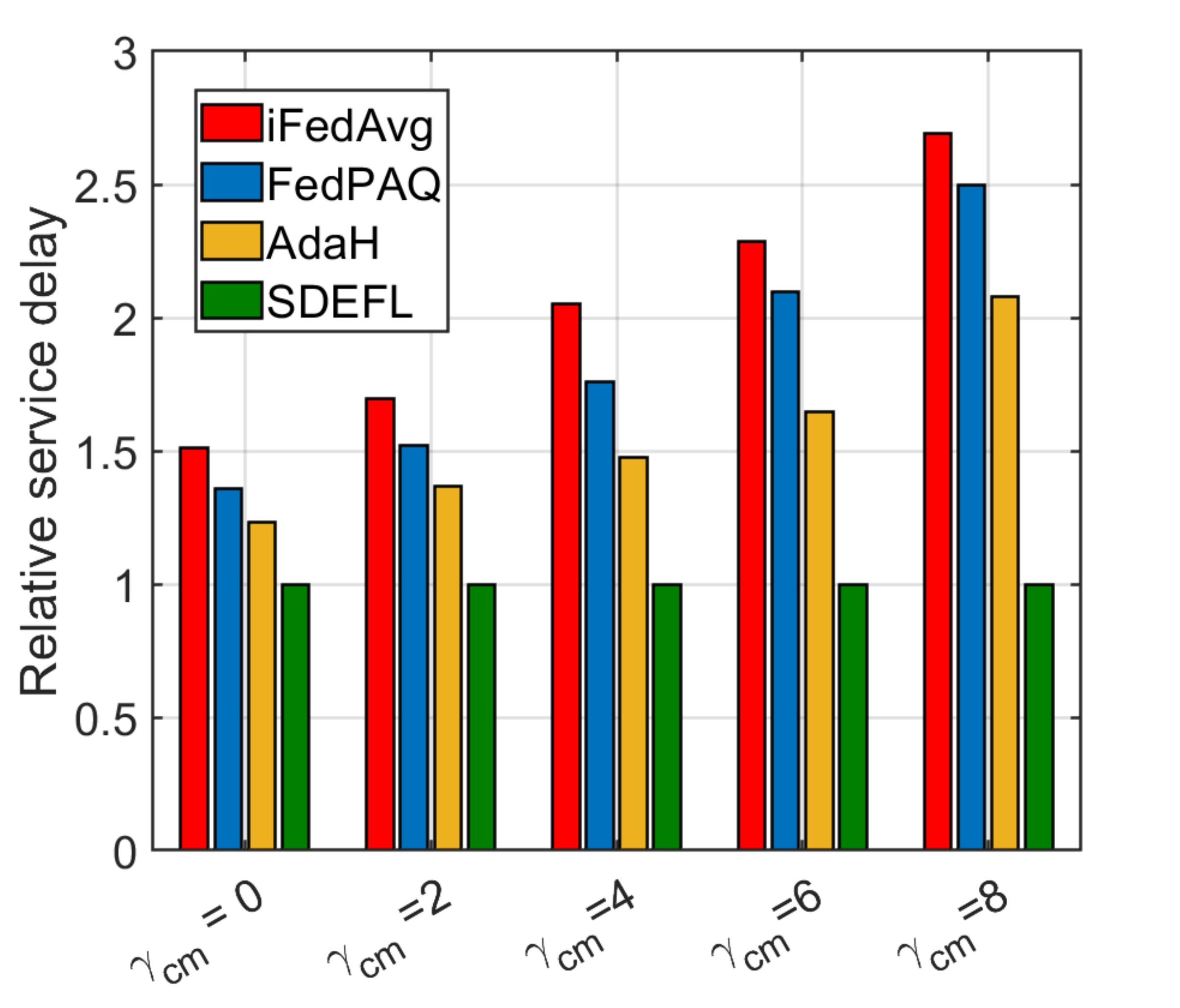}}
  %\vspace{-0.2in}
  \caption{Service delay with different heterogeneous settings.}\label{fig:heter}
  \vspace{-0.1in}
\end{figure*}

% since the model is larger relative to
\subsubsection{Impacts of Learning Model Dimensions}
We evaluate the proposed SDEFL with different learning models. In particular, the model sizes (i.e., $d$) of ResNet20 and MobileNetv2 are 0.27M and 3.4M, respectively. From Table~\ref{tab:acc}, we observe that the service delay of MobileNetv2 training is larger than that of ResNet20 since a more complex learning model requires more communication and computing delay. Besides, the proposed SDEFL can save more service delay in MobileNetv2 than that in ResNet20. In the case of MobileNetv2, it shows that when reaching FL convergence, SDEFL can reduce 46\% service delay compared with \textit{FedAvg}, reduce 38\% delay compared with \textit{FedPAQ}, and reduce 15\% delay compared with \textit{FL-QuWG-pro}. We also calculate the $\rho$ in Section~\ref{Sec:SysM1} of the system straggler: $\rho = 0.61$ in the case of ResNet and $\rho = 4.21$ in the case of MobileNetv2. The computing delay has more profound impact than communication delay in MobileNetv2. Hence, the \textit{FedPAQ} can only consume 6\% less than \textit{FedAvg}. Since SDEFL can leverage weight quantization to reduce the heavy computing workload incurred in large networks, it can effectively reduce the service delay. However, the accuracy of SDEFL drops by 6\% in MobileNetv2, which is higher than the 3\% drop in ResNet20. A larger $d$ leads to a larger loss upon convergence. Those results are consistent with the analysis in Theorems~\ref{tm:ncvx_convergence}. Besides, we apply the optimal solution derived by Eqn.~(\ref{reObj}) to perform the FL training in our testbed and measure the corresponding service delay using Pytorch. The measured delay is presented in Table~\ref{tab:acc}.
It shows that our approach has the maximum error of 12\% and an average error of 8\%. 
%We also conduct similar experiments on the devices Experimental Results (Wi-Fi 5: 88 Mbps)

\subsubsection{Impacts of Data \& Device Heterogeneity}
We further evaluate the performance of SDEFL with different data distributions in the context of skew class distribution and unbalanced number of training data samples. The model is trained by ResNet20. Sample distributions become skewer as the standard deviation (STD) goes from 0, 0.3 to 0.6. As shown in Fig.~\ref{fig:non-iid}, we find that training with non-IID data incurs a longer service delay than that of the IID case. From the results in Fig.~\ref{fig:non-iid}, we observe that a large value of $G$ requires more training iterations to converge, which verifies the analysis in Theorem~\ref{tm:ncvx_convergence}. Compared with \textit{FedAvg} The proposed SDEFL can efficiently reduced the service delay by 0.42, 0.72 and 0.6 in the case of “STD=0", “STD=0.3",  “STD=0.6", respectively.
%Note that a highly skewed data distribution has a large value of $G$ in Assumption~\ref{as:divergence}. 

We now examine the impact of device heterogeneity (DH) w.r.t computing capacities and wireless bandwidth on the service delay. Here, we divide ten participating devices into four groups, corresponding to four capacity levels. Assume that mobile devices belonging to the same group are allocated with proportional wireless bandwidth for gradient exchanges. To vary $r_n$, we use the Wi-Fi router to limit the upload bandwidth of the connected devices with a tunable parameter $\gamma_{cm}$ such as $\lambda_n -0.05\gamma_{cm}$, $\lambda_n -0.03\gamma_{cm}$, $\lambda_n + 0.03\gamma_{cm}$ and $\lambda_n + 0.05\gamma_{cm}$, respectively. The values of $\gamma_{cm} \in \{0, 2, 4, 6, 8\}$. On the other hand, to vary $t_n^{comp}$, we set $\gamma_{cp} \in \{1, 2, 3, 4\}$ to control the total types of working modes. For instance, $\gamma_{cp} = 3$ represents that mobile devices perform with three different kinds of working power and GPU cores. Higher values of $\gamma_{cm}$ and $\gamma_{cp}$ mean that devices have more diverse computing and communication conditions. In the following, we fix $\gamma_{cp}=3$ when varying $\gamma_{cm}$ and fix $\gamma_{cm}=6$ when varying $\gamma_{cp}$. From Fig.~\ref{fig:comp-heter1}, we observe that the service delay grows with the device heterogeneity, indicating a negative impact of heterogeneous mobile devices on FL training. Further, Fig.~\ref{fig:comp-heter} and Fig.~\ref{fig:comm-heter} show that our proposed scheme SDEFL can greatly reduce the service delay when the device capabilities are highly diverse since we consider the differences among mobile devices to the gradient quantization strategy and optimize the local computing control selection.
%; in the simulated environment, we set a tunable parameter $\gamma_{cm}$ as the level of heterogeneity that adjusts the bandwidth among different groups such that $\lambda_n -0.02\gamma$, $\lambda_n -0.01\gamma$, $\lambda_n + 0.01\gamma$ and $\lambda_n + 0.02\gamma$, respectively. The values of $\gamma$ range from 0 to 10. 
 
\subsubsection{ {Numerical illustration of optimal $(H, q_{g,n}, q_{w,n})$ strategies}} 
\begin{figure}[t] \centering 
\subfigure[Impact of available bandwidth to optimal $\{q_{w,n}\}$ and $H$. \label{fig:q_w}]
{\includegraphics[width=.49\linewidth]{ 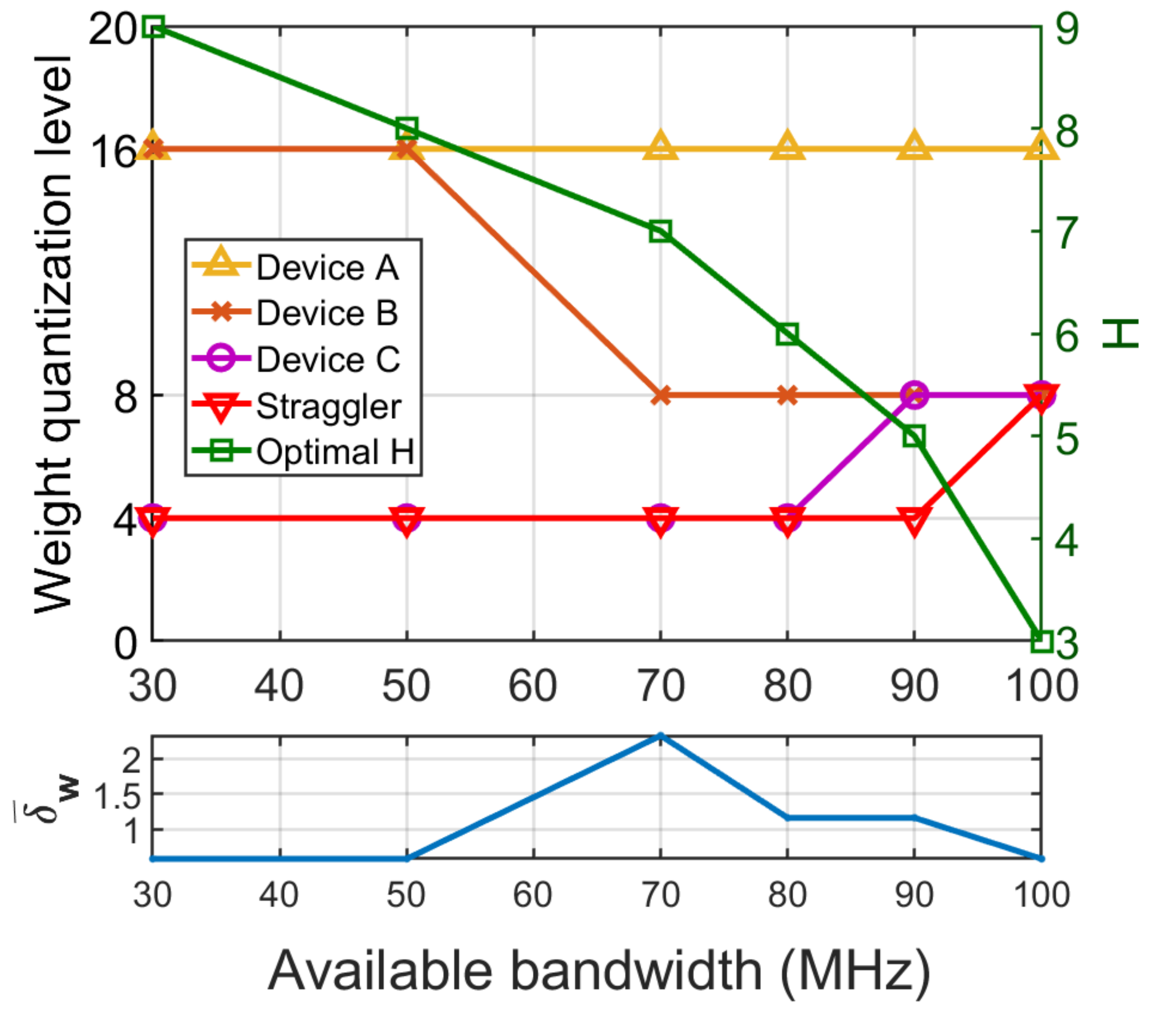}}
\subfigure[Impact of available bandwidth to optimal $\{q_{g,n}\}$ and $H$. \label{fig:q_g}]
{\includegraphics[width=.49\linewidth]{ 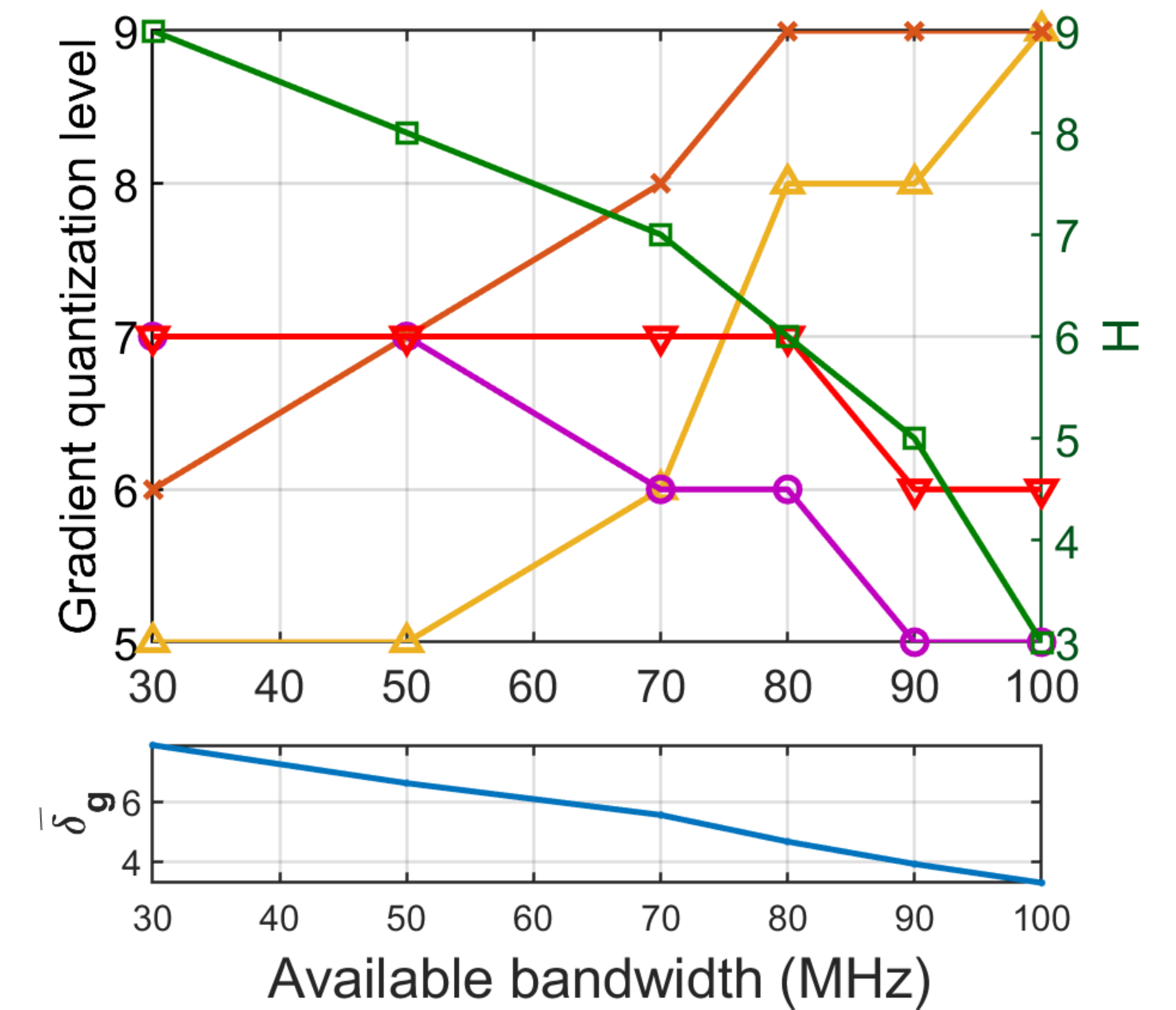}}
\caption{Optimal strategies of different devices. Here, the relationship of computing capabilities of four devices is $T^{cp}_{S} < T^{cp}_{C} < T^{cp}_{B} < T^{cp}_{A}$ and the relationship of communication conditions of four devices is $r^{cm}_{A} < r^{cm}_{B} < r^{cm}_{C} < r^{cm}_{S}$. For straggler device and Device C, they are bottlenecked by the computing delays. For Devices A and B, they are bottlenecked by the communication delays.}\label{fig:OpiSra}
 %\vspace{-0.2in}
\end{figure}

Due to the fact that the Tensor cores in Xaiver NX only support hardware acceleration for training with FP16, we only consider the feasible set of weight quantization level as $q_w \in \{ 16, 32\}$ in the testbed experiments above.

In the following, we conduct simulations with the extended feasible set of weight quantization level, and further examine the impacts of available bandwidth on the optimal values of $(H, q_{g,n}, q_{w,n})$. Here, we set the number of mobile devices 40 and the FL task is ResNet-20 with Cifar10. The expanded feasible set of weight quantization level $q_w \in \{ 4, 8, 16, 32\}$, which is a standard-setting and hardware friendly \cite{torti2018embedded}. Eqn. (\ref{Tcom}) and Eqn. (\ref{Tcomm}) are used to calculate the computing and communication delays, respectively. The optimal values of $(H, q_{g,n}, q_{w,n})$ are obtained by solving problem in (\ref{reObj1})-(\ref{reieq:t_sqrt}).

In Fig.~\ref{fig:OpiSra}, we show the optimal quantization levels of the straggler device and three other mobile devices.
%where the computing capabilities of four devices is $T^{cp}_{S} < T^{cp}_{C} < T^{cp}_{B} < T^{cp}_{A}$ and the communication conditions of four devices are $r^{cm}_{A} < r^{cm}_{B} < r^{cm}_{C} < r^{cm}_{S}$. Computing delays are a bottleneck for straggler device and device C. Communication delays are a bottleneck for devices A and B. 
We see that SDEFL prefers ``working" more and ``talking" less in the case of small wireless bandwidth, as expected. 
Furthermore, we see that the tradeoffs between ``working" and ``talking" among devices vary. Rather surprisingly, we find that, as the available bandwidth increases, for the devices with small computing capabilities, the optimal weight quantization levels $q_{w,S}^{\ast}$ and $q_{w,C}^{\ast}$ increase while their optimal gradient quantization levels decrease; for the devices with slow communication rates, for the devices with slow communication rates, the optimal weight quantization level of Device B, $q_{w,B}^{\ast}$, decreases while his optimal gradient quantization level increases. In fact, the per-round computing delay of the straggler is not increased due to a small value of $H$\footnote{In this case, the estimated per-round computing delay of ($q_{w,n}=4, H=5$) is 247ms while the estimated per-round computing delay of ($q_{w,n}=, H=5$ ) is 207ms.}. Besides, with smaller value of $\bar{\delta}_{w}$ and $\bar{\delta}_{g}$ (the bottom of Figs.~\ref{fig:q_w} and~\ref{fig:q_g}) in the case of the large available bandwidth (e.g., 90 and 100), the number of training iteration to converge (i.e.,$K$ in Eqn.~\ref{round}) is smaller. Thus the service delay can be efficiently reduced. It demonstrates that our SDEFL fully consider the device heterogeneity and allows devices to balance the workload between ``working" and ``talking" in saving their service delay during FL training. 
%As expected, when the available bandwidth is small (where the number of local training iterations $H$ is large), the straggler device is assigned with smallest weight quantization level and modest gradient quantization level; Device A is assigned with small gradient quantization level ($q_{g,A} = 4$). 

\subsubsection{Impact of the Number of Users}%
\begin{figure}[t] \centering 
  \subfigure[i.i.d distribution. \label{fig:iid_u}]
  {\includegraphics[width=.49\linewidth]{ 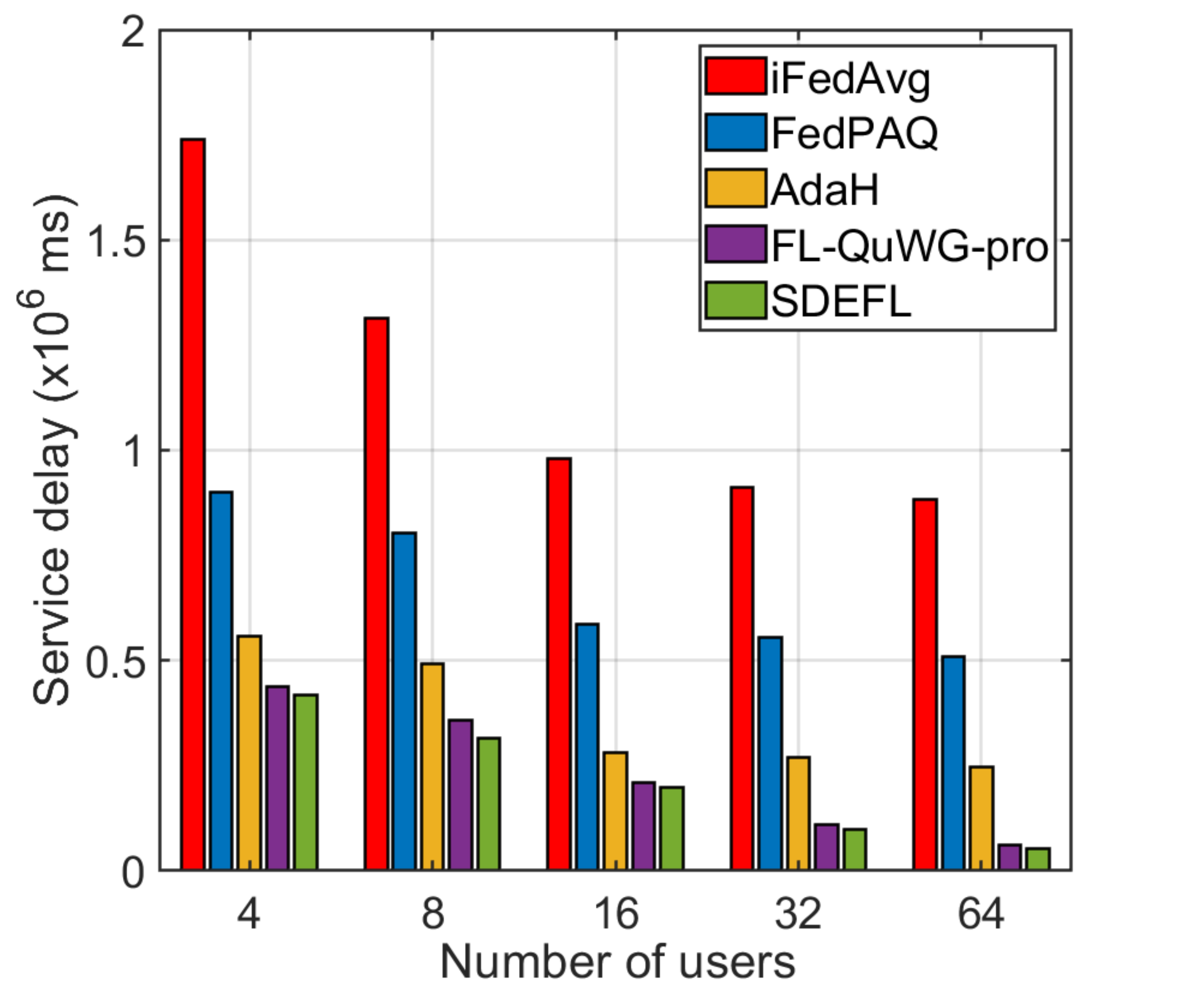}}
  \subfigure[non-i.i.d distribution. \label{fig:non-iid_u}]
  {\includegraphics[width=.49\linewidth]{ 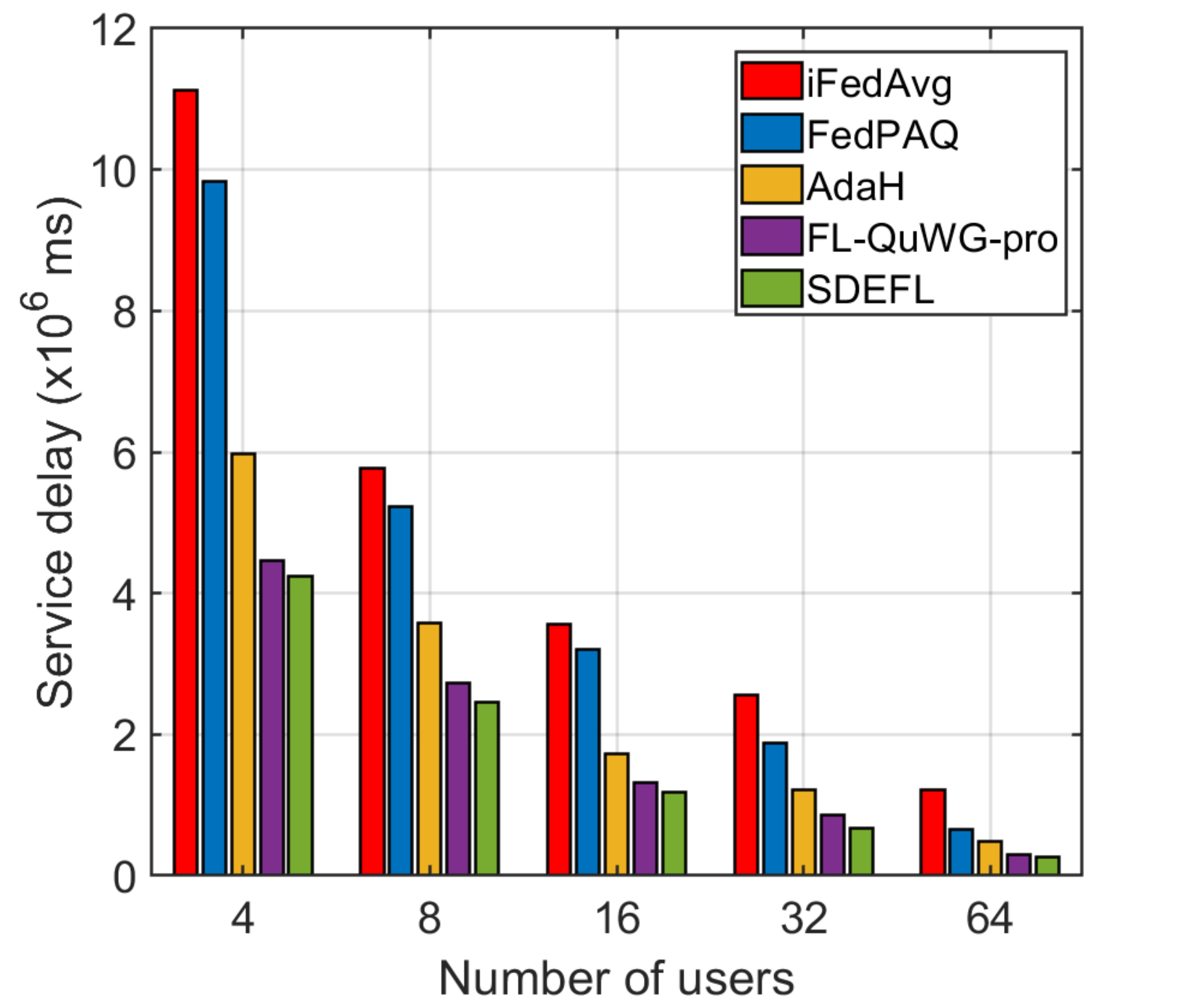}}
  \caption{Service delay with numbers of devices.}\label{fig:user}
 \vspace{-0.2in}
\end{figure}

%, which indicates the communication cost is more expensive than the computation cost
We evaluate the impact of the number of users in the simulated settings. Fig.~\ref{fig:user} demonstrates that bring more participating devices helps reduce the overall service delay in both i.i.d and non-i.i.d cases. The reason is that increasing the number of users can help speed up the convergence rate, and thereby reduce the service delay, which is also consistent with the sub-linear speedup in Theorem~\ref{tm:ncvx_convergence}. In i.i.d case, as $N$ keeps growing, the marginal service delay reduction becomes smaller and smaller. Beside, among different settings, the proposed SDEFL is superior to other schemes in terms of the service delay. 
%\vspace{-0.1in}Similar trend can be also observed in non-IID case. 

\section{Related Work}\label{sec:ReWork}
There are several research efforts made on the FL resource utilization optimization in wireless networks in both computing and communications \cite{cao2020decentralized,shi2020joint, vu2020cell, tran2019federated, luo2021cost, zhang2021federated}. Particularly, Shi et al. in \cite{shi2020joint} proposed a client scheduling scheme for FL training under targeted training delay budget. Others like Tran et al. in \cite{tran2019federated} and Vu et al. in \cite{vu2020cell} studied the tradeoffs between FL training delay and energy and conducted radio resource allocation for specific (non-optimal) design parameters, which is set based on human heuristics. Luo et al. in \cite{luo2021cost} proposed cost-efficient design and study how to determine the design parameters (i.e., $H$). However, among the previous works, the targeted learning models are either relatively simple (i.e., with convex assumptions) or shallow networks, which is inconsistent with the state-of-art DL models, and neglect to reduce the payloads from learning algorithms. A few pioneer works, such as Li et al. in \cite{li2021talk} and Shi et al. in \cite{shi2021towards} have made efforts on energy-efficient FL design on how to determine the learning parameters (e.g., sparsity ratio of model updates). However, they do not consider the mismatch between the huge computing burden on resource limited mobile devices.

Various works in machine learning literature have been developed to design efficient quantization schemes to facilitate on-device learning and communication efficient distributed learning. For on-device learning, “LQ-Net” in Zhang et al. \cite{zhang2018lq} quantized model weights such that the inner products can be computed efficiently on devices. Nevertheless, most of the weight quantization schemes only consider the case of a single device. A few works like Fu et al. \cite{fu2020don} proposed a variance-reduced weight quantization scheme in the distributed learning setting to improve the model accuracy but lacks of the discussion about its impact of how to accelerate the service delay. Different from those existing works, we study and implemented the quantization schemes on real-world mobile devices to verify their actual speedup w.r.t computing/service delay for on-device training.

For distributed learning with gradient quantization, recent theoretical works \cite{basu2019qsparse,reisizadeh2020fedpaq,alistarh2017qsgd} introduced variance-reduced methods and error-feedback schemes to improve the model prediction accuracy and their corresponding convergence analysis assume identical quantization strategies across all the mobile devices, which is not appropriate in the FL setting. Moreover, the effect of precision scaling on both model accuracy and service delay is an open problem that we aim to address in this work. Note that our proposed scheme can be easily incorporated with the resource allocation to further improve the delay efficiency.

\section{Conclusion and Future Work}\label{Sec:con}
In this paper, we have studied the service delay efficient FL (SDEFL) over mobile devices via joint design of local computing control, weight quantization and gradient quantization. We have empirically investigated the impacts of the weight quantization, gradient quantization and local computing control strategies on the service delay, and provided the convergence rate of FL with compression from a theoretical perspective. Guided by the derived theoretical convergence rate, we have investigated the tradeoff between ``working" (i.e., computing delay) and ``talking" (i.e., communication delay), and formulated the SDEFL training problem into a mixed integer programming optimization. We have converted it to the equivalent convex programming optimization, and developed feasible solutions. Extensive experimental and emulation results have demonstrated the effectiveness and efficiency of our proposed SDEFL scheme, and its advantages over peer designs in various learning and wireless transmission settings.

Note that current commercial off-the-shelf mobile Volta and Tuning GPUs can only support the speedup for FP16 training (i.e., 16-bit weight quantization). Therefore, to implement fast local model training with lower precision than 16-bit (in particular lower than 8-bit), we should jointly consider the algorithm development with the hardware/architecture design. As a possible future extension of our work, we would like to expand our existing SDEFL testbed with FPGA based Xilinx Zynq~\cite{Znyq} via hardware and software co-designs to support lower level weight quantization (e.g., 8-bit or 4-bit quantization) and show its advantages. 

%Besides, other model compression techniques such as model pruning can further reduce computing workloads and communication payload (client only need to send the pruned model updates) to minimize the service delay. 
% By integrating those compression techniques, it's worth investigating how compression parameters affect the service delay.

\appendices
\section{Proof of Theorem \ref{tm:ncvx_convergence}} \label{proof:tm} 
\subsection{Additional Notation}
For simplicity of notations, we denote the error of weight quantization ${\boldsymbol{r}}_n^{k} \triangleq Q_w \left({\boldsymbol{w}}_n^{k+1}\right) - {\boldsymbol{w}}_n^{k+1}$, and the local “gradient” with weight quantization as $\widetilde{{\boldsymbol{g}}}_n^{k} \triangleq \triangledown \widetilde{f}_n ({\boldsymbol{w}}_n^{k} ) - {\boldsymbol{r}}_n^{k}/\eta, \forall n \in \mathcal{N}$. Since the quantization scheme $Q_w$ we used is an unbiased scheme, $\mathbb{E}_Q\left[{\boldsymbol{r}}_n^{k}\right] = 0$. 

%and $\mathbb{E} \left[\mathbb{E}_Q [\hat{{\boldsymbol{g}}}_n^{k} ]\right] = \triangledown {F}_n ({\boldsymbol{w}}_n^{k})$.

Inspired by the iterate analysis framework in
we define the following virtual sequences:
{\small
\begin{flalign}
{\boldsymbol{u}}_n^{k+1} &= {\boldsymbol{w}}_n^{k} - \eta \widetilde{{\boldsymbol{g}}}_n^{k}, \\
{\boldsymbol{w}}_n^{k+1} &= 
\left\{\begin{matrix}
{\boldsymbol{u}}^{k+1}_n, & ((k+1) \mod H) \neq 0,\\ 
{\boldsymbol{u}}^{k'}_n - \sum_{n=1}^N p_n Q_{g,n}( {\Delta}_n^{k'}),& otherwise.
\end{matrix}\right. \label{update}
\end{flalign} }
Here, $k' = k+1-H$ is the last synchronization step and $\Delta^{k'}_n ={\boldsymbol{u}}^{k'}_n - {\boldsymbol{u}}^{k+1}_n$ is the differences since the last synchronization. The following short-hand notation will be found useful in the convergence analysis of the proposed FL framework:
{\small
\begin{flalign} 
     {\boldsymbol{u}}^{k} &= \sum_{n=1}^N p_n {\boldsymbol{u}}_n^{k}, \quad {\boldsymbol{w}}^{k} = \sum_{n=1}^N p_n {\boldsymbol{w}}_n^{k}, \label{n1} \\
    \widetilde{{\boldsymbol{g}}}^{k} &= \sum_{n=1}^N p_n \widetilde{{\boldsymbol{g}}}_n^{k}, \quad {{\boldsymbol{g}}}^{k}  = \sum_{n=1}^N p_n \triangledown {F}_n ({\boldsymbol{w}}_n^{k} ).\label{n2}
\end{flalign}}
Thus, ${\boldsymbol{u}}^{k+1} = {\boldsymbol{w}}^{k} - \eta \widetilde{{\boldsymbol{g}}}^{k}$. Note that we can only obtain $ {\boldsymbol{w}}^{k+1}$ when $((k+1) \mod H) = 0$. Further, due to the unbiased gradient quantization scheme, $Q_g$, no matter whether $((k+1) \mod H) = 0$ or $((k+1) \mod H) \neq 0$, we always have $\mathbb{E} [\mathbb{E}_{Q_g} [ {{\boldsymbol{w}}}^{k+1} ] ] = \mathbb{E}[  {{\boldsymbol{u}}}^{k+1}]$.

\subsection{Key Lemmas}
Now, we give four important lemmas to convey our proof.

\begin{lemma}[Bounding the weight quantization error~\cite{li2017training}]\label{l:e_r}
\begin{flalign}
     &\mathbb{E}_{Q_w}\left[\norm{\boldsymbol{r}_n^{k} }_2^2\right] \leq \eta \sqrt{d} \delta_{w,n} \tau.
\end{flalign}
\end{lemma}

\begin{lemma}\label{l2}
According to the proposed algorithm the expected inner product between stochastic gradient and full batch gradient can be bounded with:
\begin{flalign}
&\mathbb{E}  \left[\mathbb{E}_Q \left[  {\langle{ \triangledown F({\boldsymbol{w}}^{k}), {\boldsymbol{u}}^{k+1} - {\boldsymbol{w}}^{k} } \rangle} \right] \right] &\nonumber  \\
&\leq - \frac{\eta}{2} \mathbb{E}  \left[\norm{\triangledown F({\boldsymbol{w}}^{k})}_2^2\right] - \frac{\eta}{2} \mathbb{E}  \left[ \norm{ {{\boldsymbol{g}}}^{k}}_2^2\right]& \nonumber  \\
%\frac{\eta}{2} \mathbb{E}  \left[ \norm{\sum_{n=1}^N p_n \triangledown F_n({\boldsymbol{w}}_n^{k})}_2^2\right] & \nonumber  \\
&\quad+ \frac{\eta L^2}{2} \sum_{n=1}^N  p_n^2  \mathbb{E}\left[\mathbb{E}_Q\left[\norm{ {\boldsymbol{w}}^{k}-{\boldsymbol{w}}_n^{k} }_2^2\right]\right]&
\end{flalign}

\begin{proof}
\begin{flalign}
     &\mathbb{E}  \left[\mathbb{E}_Q \left[  {\left \langle{ \triangledown F({\boldsymbol{w}}^{k}),  {\boldsymbol{u}}^{k+1} - {\boldsymbol{w}}^{k} } \right \rangle} \right] \right] &\nonumber\\
     &= - \eta \mathbb{E}  \left[{ \left \langle { \triangledown F({\boldsymbol{w}}^{k}), \mathbb{E}_Q  [ \widetilde{{\boldsymbol{g}}}^{k} ]   } \right \rangle}\right] & \nonumber\\
     &\overset{(a)}{=} - \eta \mathbb{E}  \left[{ \left \langle { \triangledown F({\boldsymbol{w}}^{k}), {\boldsymbol{g}}^{k}   } \right \rangle}\right] & \nonumber\\
     &\overset{(b)}{=} \frac{\eta}{2} \mathbb{E}  \left[-\norm{\triangledown F({\boldsymbol{w}}^{k})}_2^2 - \norm{ {\boldsymbol{g}}^{k} }_2^2 + \norm{\triangledown F({\boldsymbol{w}}^{k}) - {\boldsymbol{g}}^{k} }_2^2\right]  & \nonumber\\
     %\sum_{n=1}^N p_n\triangledown F_n({\boldsymbol{w}}_n^{k}
     &\overset{(c)}{\leq} - \frac{\eta}{2} \mathbb{E}  \left[\norm{\triangledown F({\boldsymbol{w}}^{k})}_2^2\right] - \frac{\eta}{2} \mathbb{E}  \left[ \norm{ {\boldsymbol{g}}^{k} }_2^2\right] & \nonumber \\
     &\quad+ \frac{\eta L^2}{2} \sum_{n=1}^N  p_n^2  \mathbb{E}\left[\mathbb{E}_Q\left[\norm{ {\boldsymbol{w}}^{k}-{\boldsymbol{w}}_n^{k} }_2^2\right]\right] &
\end{flalign}
where $(a)$ is due to $\mathbb{E}[\widetilde{{\boldsymbol{g}}}_n^{k}]$ $= \triangledown {F}_n({{\boldsymbol{w}}}_n^{k})$ and $\mathbb{E}_Q  [ \widetilde{{\boldsymbol{g}}}^{k} ] = {\boldsymbol{g}}^{k}$ and $(b)$ is due to $-2\left<a,b\right> = \norm{a-b}^2 - \norm{a}^2 -\norm{b}^2$, and $(c)$ follows from $L$-smoothness assumption. The proof is completed.
%$(b)$ holds due to $\norm{\sum_{i=1}^{n} a_i}^2$ $\leq n\sum_{i=1}^{n}\norm{a_i}^2$, 
\end{proof}
\end{lemma}

\begin{lemma}\label{l4}
According to the proposed algorithm the expected inner product between stochastic gradient and full batch gradient can be bounded with:
\begin{flalign}
    &\frac{ L}{2} \mathbb{E} \left[\mathbb{E}_Q \left[ \norm{ {\boldsymbol{w}}^{k+1}- {\boldsymbol{w}}^{k}}_2^2 \right] \right] &\nonumber  \\
    &\leq \frac{\eta^2L \bar{p}\sigma^2 }{ 2M} + \frac{\eta^2  L}{2}  \norm{ \boldsymbol{g}^k }_2^2 + \frac{\eta L}{2} \sqrt{d} \tau \sum_{n=1}^N p_n^2 \delta_{w,n} & \nonumber\\
    &\quad +\frac{ L}{2}\sum_{n=1}^N p_n^2  \delta_{g,n} \mathbb{E} \left[||{ {\boldsymbol{u}}_n^{k}- {\boldsymbol{u}}^{k'}}||_2^2\right].
    %&\leq \left(\frac{\eta^2L\sigma^2}{2M}+ \frac{\eta L\sqrt{d} \delta_w \tau}{2} \right) (\delta_gH+1) \sum_{n=1}^N p_n^2 &\nonumber\\
    %&\quad +\frac{3\eta^2L^3\delta_gH}{2} \sum_{i=k'}^{k'_H} {A1}_i  + \frac{3\eta^2L \delta_g}{2} H \sum_{i=k'}^{k'_H} \norm{ \triangledown {F} ( {\boldsymbol{w}}^{i}) }_2^2&\nonumber\\
    %&\quad + \frac{\eta^2 L}{2}\mathbb{E} \left[\norm{{{\boldsymbol{g}}}^{k}}_2^2 \right]+ \frac{3\eta^2L}{2}H^2G^2&
    %
    %&\leq \frac{\eta^2L\sigma^2}{2M} (\bar{\delta}_g H + 1) + \frac{3\eta^2L \bar{\delta}_g}{2} H \sum_{i=k'}^{k} \norm{ \triangledown {F} (  {\boldsymbol{w}}^{i}) }_2^2&\nonumber\\
    %&\quad + \frac{\eta L}{2} \sqrt{d} \tau \sum_{n=1}^N  p_n^2 (\delta_{g,n}H+1) \delta_{w,n}  &\nonumber\\
    %&\quad + \frac{3\eta^2L^3 }{2} H \sum_{i=k'}^{k} {A1}_i + \frac{\eta^2 L}{2}\mathbb{E} \left[\norm{\sum_{n=1}^N p_n \triangledown {F}_n ({\boldsymbol{w}}_n^{k} ) }_2^2 \right].
\end{flalign}
where $k' = k+1-H$ $(k'< k)$ is the nearest synchronization step.
%where $\bar{\delta}_g = \sum_{n=1}^N p_n^2 \delta_{g,n}$.

\begin{proof}
According to the update rule defined in (\ref{update}), when $(k+1) \mod H = 0$, we have,
\begin{flalign}
    &{\boldsymbol{w}}^{k+1} - {\boldsymbol{w}}^{k} \nonumber\\
    &=  {\boldsymbol{w}}^{k+1} - {\boldsymbol{u}}^{k+1} + {\boldsymbol{u}}^{k+1} - {\boldsymbol{w}}^{k} \nonumber\\
    &={\small  \sum_{n=1}^N p_n \left({\boldsymbol{u}}^{k'}_n - \sum_{n=1}^N p_n Q_{g,n}( {\boldsymbol{u}}^{k'}_n - {\boldsymbol{u}}^{k+1}_n)\right) - {\boldsymbol{u}}^{k+1} } \nonumber\\
    &\quad + {\boldsymbol{u}}^{k+1} - {\boldsymbol{w}}^{k} \nonumber\\
    &= \sum_{n=1}^N p_n \left( {\boldsymbol{u}}^{k'}_n - {\boldsymbol{u}}^{k+1}_n - Q_{g,n}( {\boldsymbol{u}}^{k'}_n - {\boldsymbol{u}}^{k+1}_n ) + ({\boldsymbol{u}}^{k+1}_n - {\boldsymbol{w}}^{k}_n) \right)\nonumber\\
    &\overset{(a)}{=} \sum_{n=1}^N p_n \left( \widetilde{\Delta}^{k'}_{n} - Q_{g,n}( \widetilde{\Delta}^{k'}_{n}) \right) + \eta \widetilde{\boldsymbol{g}}^{k}.
\end{flalign}
Here, $(a)$ is due to the short-hand notation defined in (\ref{n1})-(\ref{n2}) and $k' = k+1-H$ is the last synchronization step. We then have,
\begin{flalign}
     &\frac{ L}{2} \mathbb{E} \left[\mathbb{E}_Q \left[ \norm{ {\boldsymbol{w}}^{k+1}- {\boldsymbol{w}}^{k}}_2^2 \right] \right] &\nonumber\\
     &= \frac{ L}{2} \mathbb{E} \left[  \mathbb{E}_Q \left[ 
     \norm{ \sum_{n=1}^N p_n  \left(  Q_{g,n}( \widetilde{\Delta}^{k'}_{n}) -\widetilde{\Delta}^{k'}_{n} \right) }_2^2
     \right] + \norm{ \eta \widetilde{\boldsymbol{g}}^{k}}_2^2 \right]&\nonumber\\
     &\overset{(a)}{\leq} \frac{\eta^2 L}{2} \mathbb{E} \left[\norm{ \widetilde{\boldsymbol{g}}^{k}}_2^2 \right]+\frac{ L}{2}\sum_{n=1}^N p_n^2  \delta_{g,n} \mathbb{E} \left[||{ {\boldsymbol{u}}_n^{k}- {\boldsymbol{u}}^{k'}}||_2^2\right] , & \label{l41} 
\end{flalign}
where $(a)$ is due to Definition~\ref{def:q}. The first term in (\ref{l41}) can be bounded by
\begin{flalign}
&\frac{\eta^2 L}{2}\mathbb{E} \left[\norm{\sum_{n=1}^N p_n( \triangledown \widetilde{f}_n ({\boldsymbol{w}}_n^{k}) -{\boldsymbol{r}}_n^{k}/\eta) }_2^2 \right]  &\nonumber\\
&=\frac{\eta^2 L}{2}\mathbb{E} \left[\norm{ \sum_{n=1}^N p_n( \triangledown \widetilde{f}_n ({\boldsymbol{w}}_n^{k}) - \triangledown {F}_n ({\boldsymbol{w}}_n^{k}) + \triangledown {F}_n ({\boldsymbol{w}}_n^{k})) }_2^2 \right]&\nonumber\\
&\quad + \frac{L}{2 } \sum_{n=1}^N p_n^2 \mathbb{E} \left[\norm{ {\boldsymbol{r}}_n^{k} }_2^2 \right]&\nonumber
\end{flalign}
\begin{flalign}
&\leq  \frac{\eta^2L \bar{p}\sigma^2 }{ 2M} + \frac{\eta^2  L}{2}  \norm{ \boldsymbol{g}^k  }_2^2 + \frac{\eta L}{2} \sqrt{d} \tau \sum_{n=1}^N p_n^2 \delta_{w,n}. \label{l431} 
\end{flalign}
Replacing the first term in (\ref{l41}) with the result in (\ref{l431}), we have 
\begin{flalign}
     &\frac{ L}{2} \mathbb{E} \left[\mathbb{E}_Q \left[ \norm{ {\boldsymbol{w}}^{k+1}- {\boldsymbol{w}}^{k}}_2^2 \right] \right] &\nonumber\\
     &\leq \frac{\eta^2L \bar{p}\sigma^2 }{ 2M} + \frac{\eta^2  L}{2}  \norm{ \boldsymbol{g}^k  }_2^2 + \frac{\eta L}{2} \sqrt{d} \tau \sum_{n=1}^N p_n^2 \delta_{w,n} \nonumber\\
     &\quad + \frac{ L}{2}\sum_{n=1}^N p_n^2  \delta_{g,n} \mathbb{E} \left[\norm{ {\boldsymbol{u}}_n^{k}- {\boldsymbol{u}}^{k'}}_2^2\right].
     %&\leq \frac{\eta^2L\sigma^2}{2M} (\bar{\delta}_g H + 1) + \frac{3\eta^2L \bar{\delta}_g}{2} H \sum_{i=k'}^{k'_H} \norm{ \triangledown {F} (  {\boldsymbol{w}}^{i}) }_2^2&\nonumber\\
     %&\quad + \frac{\eta L}{2} \sqrt{d} \tau \sum_{n=1}^N  p_n^2 (\delta_{g,n}H+1) \delta_{w,n}  &\nonumber\\
     %&\quad + \frac{3\eta^2L^3 }{2} H \sum_{i=k'}^{k'_H} {A1}_i + \frac{\eta^2 L}{2}\mathbb{E} \left[\norm{\sum_{n=1}^N p_n \triangledown {F}_n ({\boldsymbol{w}}_n^{k} ) }_2^2 \right].
\end{flalign}
The proof is completed.
%By combining the results in (\ref{l432}) and (\ref{l431}), we have 
%\begin{flalign}
%     &\frac{ L}{2} \mathbb{E} \left[\mathbb{E}_Q \left[ \norm{ {\boldsymbol{w}}^{k+1}- {\boldsymbol{w}}^{k}}_2^2 \right] \right] &\nonumber\\
     %
%     &\leq \frac{\eta^2L\sigma^2}{2M} (\bar{\delta}_g H + 1) + \frac{3\eta^2L \bar{\delta}_g}{2} H \sum_{i=k'}^{k'_H} \norm{ \triangledown {F} (  {\boldsymbol{w}}^{i}) }_2^2&\nonumber\\
%    &\quad + \frac{\eta L}{2} \sqrt{d} \tau \sum_{n=1}^N  p_n^2 (\delta_{g,n}H+1) \delta_{w,n}  &\nonumber\\
%     &\quad + \frac{3\eta^2L^3 }{2} H \sum_{i=k'}^{k'_H} {A1}_i + \frac{\eta^2 L}{2}\mathbb{E} \left[\norm{\sum_{n=1}^N p_n \triangledown {F}_n ({\boldsymbol{w}}_n^{k} ) }_2^2 \right].
%\end{flalign}
\end{proof}
\end{lemma}

\begin{lemma}[Bounding the divergence] \label{le:3}
Suppose $1-3\eta^2 L^2 H^2 > 0$, we have,
\begin{flalign}
     &\sum_{k=0}^{K-1} \sum_{n=1}^N p_n^2  \mathbb{E}\left[\mathbb{E}_Q\left[\norm{  {\boldsymbol{w}}^{k}-{\boldsymbol{w}}_n^{k} }_2^2\right]\right] &\nonumber\\
     &\frac{ \eta^2 KH \bar{p}\sigma^2}{M(1-3\eta^2 L^2 H^2)} + \frac{\eta \sqrt{d} KH \tau \sum_{n=1}^N  p_n^2 \delta_{w,n} }{1-3\eta^2 L^2 H^2 }& \nonumber\\
     &\quad + \frac{3\eta^2 H^2 \bar{p}}{1-3\eta^2 L^2 H^2 }\sum_{i=0}^{K-1}\norm{ \triangledown {F} ( {\boldsymbol{w}}^{i}) }_2^2, &
\end{flalign}
and 
\begin{flalign}
&\sum_{n=1}^N p_n^2  \delta_{g,n} \mathbb{E} \left[ \norm{ {\boldsymbol{u}}_n^{k}- {\boldsymbol{u}}^{k'}}_2^2\right] &  \nonumber\\
&\leq \eta^2 H \frac{\bar{\delta}_g \sigma^2}{M} + 3\eta^2 \bar{\delta}_g H \sum_{i=k'}^{k'_H}  \norm{ \triangledown {F} ( {\boldsymbol{w}}^{i}) }_2^2 &  \nonumber\\
&\quad +  3\eta^2 L^2 H \sum_{i=k'}^{k'_H} A1_k + \eta \sqrt{d} H \sum_{n=1}^N  p_n^2 \delta_{g,n}\delta_{w,n} \tau.
\end{flalign}
where $\bar{p} = \sum_{n=1}^N p_n^2$ and $\bar{\delta}_g = \sum_{n=1}^N p_n^2 \delta_{g,n}$.

\begin{proof}
Recalling that at the synchronization step where ($k' \mod H = 0$), $\boldsymbol{w}_n^{k'}= {\boldsymbol{w}}^{k'}$ for all $n \in \mathcal{N}$. Therefore, for any $k \geq 0$, such that $k' \leq k \leq k'+H$, we get,
\begin{flalign}
    A1_k
    &:=\sum_{n=1}^N p_n^2   \mathbb{E}\left[\mathbb{E}_Q\left[\norm{  {\boldsymbol{w}}^{k} - {\boldsymbol{w}}_n^{k} }_2^2\right]\right] \nonumber\\
    &= \sum_{n=1}^N p_n^2  \mathbb{E}\left[\mathbb{E}_Q\left[ \norm{ ( {\boldsymbol{w}}^{k} - {\boldsymbol{w}}^{k'}) - ({\boldsymbol{w}}_n^{k}- {\boldsymbol{w}}^{k'}) }_2^2\right]\right]\nonumber\\
    &\overset{(a)}{\leq} \sum_{n=1}^N p_n^2  \mathbb{E}\left[\mathbb{E}_Q\left[ \norm{ {\boldsymbol{w}}_n^{k}- {\boldsymbol{w}}_n^{k'} }_2^2 \right]\right]\nonumber \\
    &= \sum_{n=1}^N p_n^2  \mathbb{E}\left[\norm{\sum_{i=k'}^{k} \eta \triangledown \widetilde{f}_n ({\boldsymbol{w}}_n^{i} ) - \sum_{i=k'}^{k} \boldsymbol{r}_n^i  }_2^2\right]\nonumber
    \end{flalign}
\begin{flalign}
    &\overset{(b)}{=} \eta^2 \sum_{n=1}^N p_n^2  \mathbb{E}\left[\norm{\sum_{i=k'}^{k}\triangledown \widetilde{f}_n ({\boldsymbol{w}}_n^{i} )}_2^2\right] \nonumber\\
    &\quad +  \sum_{n=1}^N p_n^2 \sum_{i=k'}^{k}\mathbb{E}_Q\left[\norm{ \boldsymbol{r}_n^i }_2^2\right] \nonumber\\
    &\leq \eta^2 \sum_{n=1}^N p_n^2  \mathbb{E}\left[\norm{\sum_{i=k'}^{k'_H} \triangledown \widetilde{f}_n ({\boldsymbol{w}}_n^{i} ) }_2^2 \right] \nonumber\\
    &\quad + H \eta \sqrt{d} \sum_{n=1}^N  p_n^2 \delta_{w,n} \tau, \label{eq:n3}  
\end{flalign}
where $k'_H = k'+H-1$, $(a)$ results from $\sum_{n=1}^{N} p_n ({\boldsymbol{w}}_i^{k}- {\boldsymbol{w}}_i^{k'}) = {\boldsymbol{w}}^{k} - {\boldsymbol{w}}^{k'}$, ${\boldsymbol{w}}_i^{k'} =  {\boldsymbol{w}}^{r'}$, and $\mathbb{E} \lVert{\boldsymbol{x} - \mathbb{E}[\boldsymbol{x}]} \rVert_2^2 \leq \mathbb{E}\lVert{\boldsymbol{x} }\rVert_2^2$. $(b)$ holds due to the unbiased quantization scheme, i.e., $\mathbb{E}_Q\left[{\boldsymbol{r}}_n^{k}\right] = 0$.

We generalize the result from \cite{jiang2018linear} to upper-bound the first term in RHS of (\ref{eq:n3}),  (see the of Theorem 3 and its proof in appendix for the special case of $p_n = \frac{1}{N}$):
\begin{flalign}
     &\eta^2 \sum_{n=1}^N p_n^2  \mathbb{E}\left[\norm{\sum_{i=k'}^{k'_H} (  \triangledown \widetilde{f}_n ({\boldsymbol{w}}_n^{i} )- \triangledown{F}_n ({\boldsymbol{w}}_n^{i} )+ \triangledown{F}_n ({\boldsymbol{w}}_n^{i} ) ) }_2^2\right] \nonumber\\
     %
     %&\leq \eta^2 H \frac{\bar{p} \sigma^2}{M}  + 3\eta^2 H^2 G^2 + 3\eta^2 L^2 H \sum_{i=k'}^{k'_H} {A1}_i \nonumber\\
     %&\quad  + 3\eta^2  H \sum_{i=k'}^{k'_H} \sum_{n=1}^N p_n^2  \norm{ \triangledown {F} ( {\boldsymbol{w}}^{i}) }_2^2. \nonumber\\
     %
     &\leq \eta^2 H \frac{\bar{p} \sigma^2}{M} + 3\eta^2 L^2 H \sum_{i=k'}^{k'_H} \sum_{n=1}^N p_n^2 \norm{  {\boldsymbol{w}}^{i} - {\boldsymbol{w}}_n^{i}}_2^2 \nonumber\\
     &\quad + 3\eta^2 \bar{p} H^2 G^2 + 3\eta^2 \bar{p} H \sum_{i=k'}^{k'_H}  \norm{ \triangledown {F} ( {\boldsymbol{w}}^{i}) }_2^2,
     %
     %&\leq 3\eta^2  H \sum_{i=1}^N o_i^2 \sum_{j=r'}^{r'_H} \left( L^2 \norm{ \bar{\boldsymbol{w}}^{j}- {\boldsymbol{w}}_i^{j}}_2^2  +  \norm{ \triangledown {F} ( \bar{\boldsymbol{w}}^{j}) }_2^2\right)   
\end{flalign}
where $\bar{p} = \sum_{n=1}^N p_n^2$. It follows that
\begin{flalign}
     \sum_{i=0}^{K-1} A1_i  
     &\leq \eta^2 KH\bar{p}(\frac{\sigma^2}{M} + 3HG^2)
      + KH \eta \sqrt{d} \tau \sum_{n=1}^N  p_n^2 \delta_{w,n}  & \nonumber\\
      %\eta^2 KH \frac{\bar{p}\sigma^2}{M} + 3\eta^2 \bar{p} KH^2 G^2
     %&\quad + 3\eta^2 L^2 H^2 \sum_{i=0}^{K-1} {A1}_i + 3\eta^2 \bar{p} H^2 \sum_{i=0}^{K-1} \norm{ \triangledown {F} (  {\boldsymbol{w}}^{i}) }_2^2 .
     &\quad + 3 \eta^2 H^2 \sum_{i=0}^{K-1} \left( L^2 {A1}_i + \bar{p} \norm{ \triangledown {F} (  {\boldsymbol{w}}^{i}) }_2^2 \right). &
\end{flalign}

Suppose $1-3\eta^2 L^2 H^2 > 0$, we have
\begin{flalign}
    \sum_{i=0}^{K-1} A1_i  
    &\leq \frac{ \eta^2 KH \bar{p}(\sigma^2/M+3HG^2)}{(1-3\eta^2 L^2 H^2)} & \nonumber\\
    &\quad \frac{\eta \sqrt{d} KH \tau \sum_{n=1}^N  p_n^2 \delta_{w,n} }{1-3\eta^2 L^2 H^2 }& \nonumber\\
    &\quad + \frac{3\eta^2 H^2 \bar{p}}{1-3\eta^2 L^2 H^2 }\sum_{i=0}^{K-1}\norm{ \triangledown {F} ( {\boldsymbol{w}}^{i}) }_2^2,
\end{flalign}

%When $(k+1) \mod H = 0$, the second term in (\ref{l41}) can be bounded by

Similarly, we denote 
\begin{equation}
    A2_k :=  \sum_{n=1}^N p_n^2  \delta_{g,n} \mathbb{E} \left[ \norm{ {\boldsymbol{u}}_n^{k}- {\boldsymbol{u}}^{k'}}_2^2\right],
\end{equation}
and have
\begin{flalign}
    A2_k & \leq  \eta^2 \sum_{n=1}^N p_n^2 \delta_{g,n} \mathbb{E}\left[\norm{\sum_{i=k'}^{k'_H} \triangledown \widetilde{f}_n ({\boldsymbol{w}}_n^{i} ) }_2^2 \right]& \nonumber\\
    &\quad + \eta \sqrt{d} H \sum_{n=1}^N  p_n^2 \delta_{g,n}\delta_{w,n} \tau &\nonumber\\
    &\leq \eta^2 H \frac{\bar{\delta}_g \sigma^2}{M} + 3\eta^2 \bar{\delta}_g H \sum_{i=k'}^{k'_H}  \norm{ \triangledown {F} ( {\boldsymbol{w}}^{i}) }_2^2 &  \nonumber\\
    &\quad +  3\eta^2 L^2 H \sum_{i=k'}^{k'_H} A1_k + \eta \sqrt{d} H \sum_{n=1}^N  p_n^2 \delta_{g,n}\delta_{w,n} \tau. &
    % +  \frac{3\eta^2 L^3}{2} H \sum_{i=k'}^{k'_H} A1_i 
    %&\quad  +  3\eta^2 L^2 H \sum_{i=k'}^{k'_H}  \sum_{n=1}^N p_n^2  \delta_{g,n} \mathbb{E} \left[\norm{ {\boldsymbol{u}}_n^{k}- {\boldsymbol{u}}^{k'}}_2^2\right]\nonumber\\
\end{flalign}
and the proof is completed.
%It follows the result from (\ref{eq:n3}). Here, $\bar{\delta}_g = \sum_{n=1}^N p_n^2 \delta_{g,n}$. We then have
%\begin{flalign}
%\sum_{k=0}^{K-1} A2_k & \leq \frac{\eta^2 \sigma^2}{M} KH\bar{\delta}_g + \eta \sqrt{d} KH \sum_{n=1}^N  p_n^2 \delta_{g,n}\delta_{w,n} \tau & \nonumber\\
%&\quad + 3\eta^2H \sum_{k=0}^{K-1} \left(L^2 A2_k + \bar{\delta}_g \norm{ \triangledown {F} ( {\boldsymbol{w}}^{k}) }_2^2 \right).
%
%( 1- 3\eta^2L^2H) \sum_{k=0}^{K-1} A2_k & \nonumber\\
%&\leq \frac{\eta^2 \sigma^2}{M} KH\bar{\delta}_g + 3\eta^2 H\bar{\delta}_g \sum_{k=0}^{K-1} \norm{ \triangledown {F} ( {\boldsymbol{w}}^{k}) }_2^2 &\nonumber\\
%&\quad + \eta \sqrt{d} H \sum_{n=1}^N  p_n^2 \delta_{g,n}\delta_{w,n} \tau. &  
%\end{flalign}
%Suppose $1-3\eta^2 L^2 H > 0$, we have
%{\small
%\begin{flalign}
%\sum_{k=0}^{K-1} A2_k 
%
%&\leq \frac{\eta^2 \sigma^2KH\bar{\delta}_g}{ M( 1- 3\eta^2L^2H)} + \frac{3\eta^2 H\bar{\delta}_g \sum_{k=0}^{K-1} \norm{ \triangledown {F} ( {\boldsymbol{w}}^{k}) }_2^2}{1- 3\eta^2L^2H} &\nonumber\\
%&\quad + \frac{\eta  \sqrt{d} H \sum_{n=1}^N  p_n^2 \delta_{g,n}\delta_{w,n} \tau}{1- 3\eta^2L^2H}, & 
%\end{flalign}
%}

\end{proof}
\end{lemma}

\subsection{Main Results}
Under the $L$-smooth assumption of $F$, we have,
\begin{flalign}
     &\mathbb{E} \left[ F( {\boldsymbol{w}}^{k+1}) - F( {\boldsymbol{w}}^{k}) \right]  & \nonumber\\
     &\leq \mathbb{E} \left[  {\langle  {\triangledown F( {\boldsymbol{w}}^{k}),  {\boldsymbol{w}}^{k+1}- {\boldsymbol{w}}^{k}} \rangle} \right] + \frac{L}{2} \mathbb{E} \left[ \norm{ {\boldsymbol{w}}^{k+1}- {\boldsymbol{w}}^{k}}_2^2 \right] & \nonumber\\
     &= \mathbb{E}  \left[{\langle  {\triangledown F( {\boldsymbol{w}}^{k}),  {\boldsymbol{w}}^{k+1}- {\boldsymbol{u}}^{k+1} +  {\boldsymbol{u}}^{k+1}- {\boldsymbol{w}}^{k}} \rangle}\right] \nonumber\\
     &\quad+ \frac{ L}{2} \mathbb{E} \left[\norm{ {\boldsymbol{w}}^{k+1}- {\boldsymbol{w}}^{k}}_2^2 \right]&  \nonumber\\ 
     &{\leq} \mathbb{E} [{\langle{ \triangledown F( {\boldsymbol{w}}^{k}),   {\boldsymbol{u}}^{k+1} -  {\boldsymbol{w}}^{k} } \rangle}  + \frac{L}{2} \norm{ {\boldsymbol{w}}^{k+1} - {\boldsymbol{w}}^{k}}_2^2 ].&  \label{eq:n1} 
\end{flalign}
We use Lemma \ref{l:e_r}-\ref{le:3} to upper bound the RHS of (\ref{eq:n1}) and set $\eta L\leq 1$, which gets,
\begin{flalign}
    &\mathbb{E} \left[ F( {\boldsymbol{w}}^{k+1}) - F( {\boldsymbol{w}}^{k}) \right] &\nonumber\\
    &\leq - \frac{\eta}{2} \mathbb{E}  \left[\norm{\triangledown F( {\boldsymbol{w}}^{k})}_2^2\right]  + \frac{\eta L^2}{2} A1_k +\frac{ L}{2} A2_k + \frac{\eta^2L \sigma^2 }{ 2M}  & \nonumber\\
    &\quad + \frac{\eta L}{2} \sqrt{d} \tau \sum_{n=1}^N p_n^2 \delta_{w,n} - \frac{\eta}{2}( 1-\eta L) \norm{ \boldsymbol{g}^k}_2^2 & \nonumber\\
    &\overset{(a)}{\leq} - \frac{\eta}{2} \mathbb{E}  \left[\norm{\triangledown F( {\boldsymbol{w}}^{k})}_2^2\right]  + \frac{\eta L^2}{2} A1_k +\frac{ L}{2} A2_k + \frac{\eta^2L \sigma^2 }{ 2M}  & \nonumber\\
    &\quad + \frac{\eta L}{2} \sqrt{d} \tau \sum_{n=1}^N p_n^2 \delta_{w,n} \nonumber\\
    &\overset{(b)}{\leq} - \frac{\eta}{2} \mathbb{E}  \left[\norm{\triangledown F( {\boldsymbol{w}}^{k})}_2^2\right] + \frac{3\eta^2L\bar{\delta}_g}{2} H \sum_{i=k'}^{k'_H} \mathbb{E}  \left[\norm{\triangledown F( {\boldsymbol{w}}^{i})}_2^2\right]& \nonumber\\
    &\quad + \frac{\eta^2L\sigma^2}{2M} (\bar{\delta}_gH+\bar{p}) + \frac{\eta L \sqrt{d}\tau}{2} H\sum_{n=1}^N p_n^2  (\delta_{g,n}+1) \delta_{w,n}\nonumber\\
    &\quad + \frac{\eta L^2}{2} A1_k + 3\eta^2L^3H\sum_{i=k'}^{k'_H}A1_k.
\end{flalign}
where $(a)$ holds due to the setting, i.e., $\eta L \leq 1$, and $(b)$ follows the second result in Lemma~\ref{le:3}.

Summing up for all $K$ communication rounds and rearranging the terms gives,
{\small
\begin{flalign}
    &\mathbb{E} \left[ F( {\boldsymbol{w}}^{K}) - F( {\boldsymbol{w}}^{0}) \right] & \nonumber \\
    &\leq -\frac{\eta C_1}{2}  \sum_{k=0}^{K-1} \mathbb{E}  \left[\norm{\triangledown F( {\boldsymbol{w}}^{k})}_2^2\right] + \frac{\eta L^2C_2}{2}\sum_{k=0}^{K-1} A1_k & \nonumber\\
    &\quad + \frac{\eta^2L\sigma^2}{2M} (\bar{\delta}_gH+\bar{p}) + \frac{\eta L \sqrt{d}\tau}{2} H\sum_{n=1}^N p_n^2  (\delta_{g,n}+1) \delta_{w,n}, &\label{eq:n4}
    %&\leq -\frac{\eta}{2}  \sum_{k=0}^{K-1} \mathbb{E}  \left[\norm{\triangledown F( {\boldsymbol{w}}^{k})}_2^2\right] + \frac{\eta L^2 }{2}\sum_{k=0}^{K-1} A1_k   & \nonumber \\
    %&\quad + \frac{L}{2} \sum_{k=0}^{K-1} A2_k + \frac{\eta^2L K \sigma^2 }{2M} + \frac{\eta L}{2} K \sqrt{d} \tau \sum_{n=1}^N p_n^2 \delta_{w,n}& \nonumber \\
    %
    %&\leq -\frac{\eta C_1}{2}  \sum_{k=0}^{K-1} \mathbb{E}  \left[\norm{\triangledown F( {\boldsymbol{w}}^{k})}_2^2\right] + \frac{\eta^2 L C_2 KH\sigma^2}{M(1-3\eta^2 L^2 H^2)}  & \nonumber \\
    %&\quad + \frac{\eta^2 L C_2 KH\sigma^2}{M(1-3\eta^2 L^2 H^2)}  & \nonumber \\
    %
\end{flalign}
}
where $C_1=1- 3\eta L H\bar{\delta}_g$ and $C_2=1+ 3\eta L H $.
%$C_1 = 1 - \frac{3 \eta H^2 (\bar{p} + \bar{\delta}_g)}{1-3\eta^2 L^2 H^2}$ and $C_2= L\bar{p} +\bar{\delta}_g$. 

Plugging Lemma \ref{le:3} into (\ref{eq:n4}), if $C_1' = C_1 - \frac{3\eta^2H^2\bar{p}(L^2 + 3\eta L^3 H)}{1-3\eta^2 L^2 H^2} \geq 0$, we have,
\begin{flalign}
    & \frac{1}{K}  \sum_{k=0}^{K-1} \mathbb{E}  \left[\norm{\triangledown F( {\boldsymbol{w}}^{k})}_2^2\right]& \nonumber \\
    &\leq \frac{2\mathbb{E} \left[ F( {\boldsymbol{w}}^{0}) - F( {\boldsymbol{w}}^{K}) \right] }{\eta C_1' K} + \frac{\eta^2 L^2C_2 H\bar{p}\sigma^2}{  C_1'M(1-3\eta^2 L^2 H^2)}& \nonumber \\
    & \quad 
    +\frac{\eta  L^2C_2 H \sqrt{d} \tau \bar{\delta}_w}{  C_1'(1-3\eta^2 L^2 H^2)} + \frac{  L \sqrt{d} \tau H}{ C_1'} \sum_{n=1}^N p_n^2  (\delta_{g,n}+1) \delta_{w,n} & \nonumber \\
    &\quad +\frac{3\eta^2 L^2C_2 H^2 G^2}{  C_1'(1-3\eta^2 L^2 H^2)} + \frac{ \eta L  \sigma^2 (\bar{\delta}_gH+\bar{p}) }{ C_1'M},
\end{flalign}
where $\bar{\delta}_w = u \sum_{n=1}^N p_n^2 \delta_{w,n}$.

If we set $\eta = \sqrt{MN/K}$ and
\begin{equation}
    \eta L H\delta_g \geq \frac{ \eta^2H^2\bar{p}(L^2+3\eta L^3 H)}{1-3\eta^2 L^2 H^2},
\end{equation} 
we can get the $1/C'_1 \leq 2$. Thus,
\begin{flalign}
    & \frac{1}{K}  \sum_{k=0}^{K-1} \mathbb{E}  \left[\norm{\triangledown F( {\boldsymbol{w}}^{k})}_2^2\right]& \nonumber \\
    &\leq \frac{4\mathbb{E} \left[ F( {\boldsymbol{w}}^{0}) - F( {\boldsymbol{w}}^{K}) \right] }{\sqrt{MNK}} + \frac{ 2 L \sigma^2 (2H\bar{\delta}_g+p) }{\sqrt{MNK}}& \nonumber \\
    & \quad
    +\frac{12M L H \bar{\delta}_g G^2}{ \sqrt{MNK} }  + 2L\sqrt{d}\tau \sum_{n=1}^N p_n^2 (\delta_{g,n} +1)  \delta_{w,n}.
    %(\delta_gH+\delta_g+1)+ \frac{2 L \delta_g \sigma^2}{\sqrt{MK}}
    %
\end{flalign}
and the proof is completed.

\bibliographystyle{IEEEtran}  
\bibliography{SDEFL_main}

\end{document}